\pdfoutput=1
\documentclass[11pt]{article}

\usepackage[final]{acl}
\usepackage{times}
\usepackage{latexsym}

\usepackage[T1]{fontenc}
\usepackage[utf8]{inputenc}

\usepackage{microtype}

\usepackage{inconsolata}

\usepackage{graphicx}
\usepackage{algorithm}
\usepackage{algpseudocode}
\usepackage{booktabs}
\usepackage{wrapfig}
\usepackage{multirow}
\usepackage{tabularx}
\usepackage{amsmath}
\usepackage{amssymb}
\usepackage{mathtools}
\usepackage{amsthm}
\usepackage{caption}
\usepackage{colortbl}
\usepackage{comment}
\usepackage{paralist}
\usepackage{tcolorbox}
\usepackage{ragged2e}
\tcbuselibrary{listings}
\newtcolorbox[auto counter, number within=section]{rqbox}[1][]{colback=gray!9!white, colframe=black, 
    boxrule=0.3mm, 
    arc=3mm, 
    auto outer arc,
    boxsep=0mm,
    #1
}
\usepackage{placeins} 
\definecolor{purple}{RGB}{128,0,128}

\title{Not All Adapters Matter: Selective Adapter Freezing for Memory-Efficient
Fine-Tuning of Language Models}

\usepackage{authblk}
\author[1]{Hyegang Son\thanks{Equal contribution.}}
\author[1]{Yonglak Son\textsuperscript{*}}
\author[2,3]{Changhoon Kim\thanks{Work completed as part of Ph.D. research at ASU.}\thanks{Corresponding authors.}}
\author[1]{Young Geun Kim\textsuperscript{\ddag}}
\affil[1]{Korea University, \texttt{\{hyegang\_son, yonglak\_son, younggeun\_kim\}@korea.ac.kr}}
\affil[2]{Arizona State University, \texttt{kch@asu.edu}}
\affil[3]{Soongsil University, \texttt{kch@ssu.ac.kr}}

\begin{document}

\maketitle

\begin{abstract}
Transformer-based large-scale pre-trained models achieve great success. Fine-tuning is the standard practice for leveraging these models in downstream tasks. Among the fine-tuning methods, adapter-tuning provides a parameter-efficient fine-tuning by introducing lightweight trainable modules while keeping most pre-trained parameters frozen. However, existing adapter-tuning methods still impose substantial resource usage. Through our investigation, we show that each adapter unequally contributes to both task performance and resource usage. Motivated by this insight, we propose Selective Adapter FrEezing (SAFE), which gradually freezes less important adapters early to reduce unnecessary resource usage while maintaining performance. In our experiments, SAFE reduces memory usage, computation amount, and training time by 42.85\%, 34.59\%, and 11.82\%, respectively, while achieving comparable or better task performance compared to the baseline. We also demonstrate that SAFE induces regularization effect, thereby smoothing the loss landscape, which enables the model to generalize better by avoiding sharp minima.
\end{abstract}

%\input{Abstract}
%%%%%%%%%%%%%%%%%%%%%%%%%%%%%%%%%%%%%%%%%%%%%%%%%%%%%%%%%
% INTRODUCTION
%%%%%%%%%%%%%%%%%%%%%%%%%%%%%%%%%%%%%%%%%%%%%%%%%%%%%%%%%
\section{Introduction} \label{Section 1}
Large-scale pre-trained language models (PLMs) have manifested superior performance in various tasks \cite{kenton2019bert, liu2019roberta, radford2019language, yang2019xlnet}. However, training PLMs from the scratch is time-consuming and resource-intensive. Common practice has been hence to fine-tune the large-scale pre-trained models by adapting all the parameters with the downstream tasks, i.e., full parameter fine-tuning (full-tuning). 

\begin{figure}[t]
\begin{center}
\centerline{\includegraphics[width=\columnwidth]{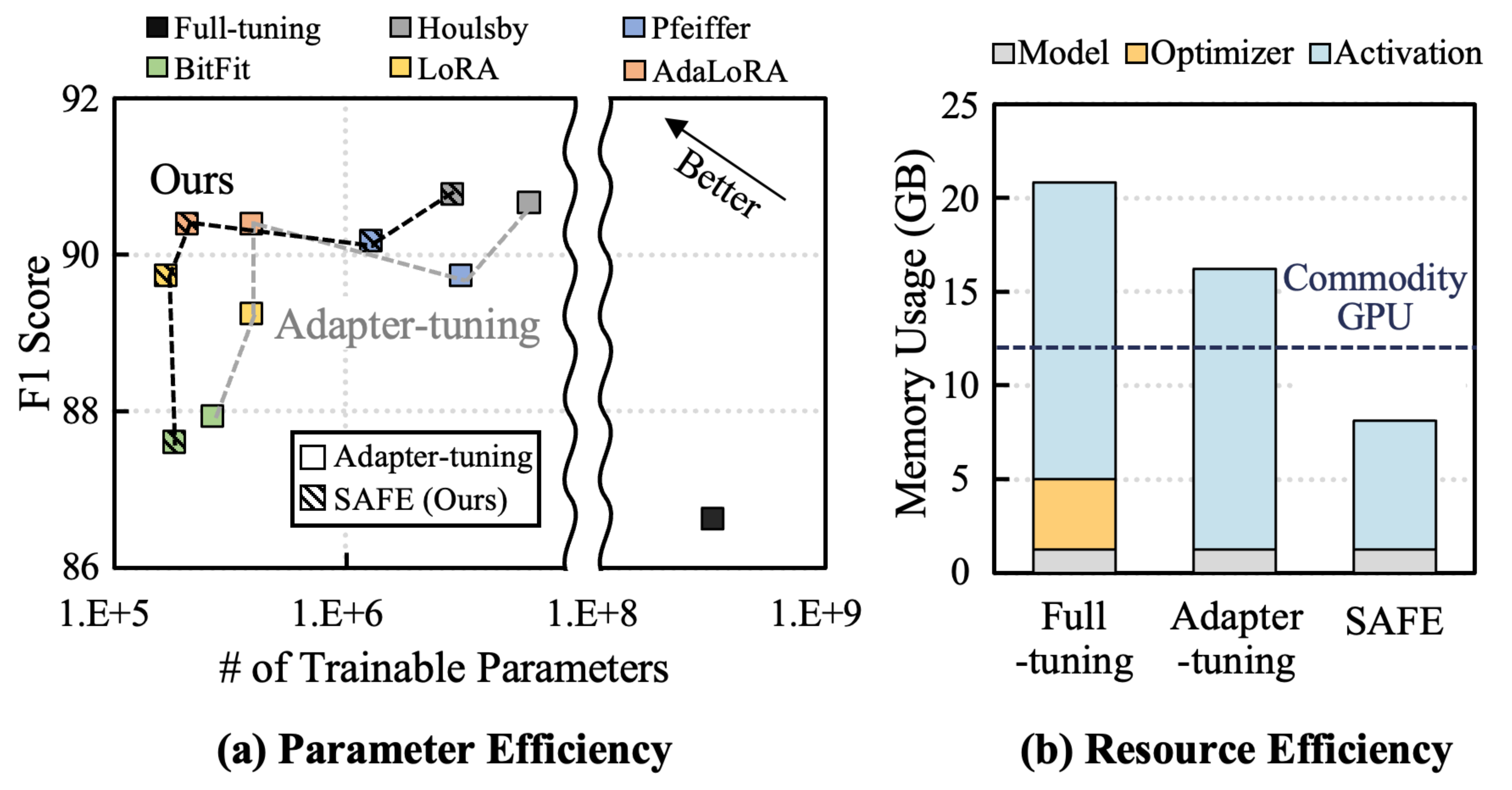}}
\caption{Comparison between full-parameter fine-tuning, adapter-tuning and our proposed SAFE on the $\text{BERT}_\text{large}$ model with SQuAD dataset. SAFE significantly reduces memory usage while providing comparable accuracy to adapter-tuning.}
\label{fig_1}
\end{center}
\vskip -0.4in
\end{figure}

Recently, Parameter-Efficient Fine-Tuning (PEFT), which focuses on optimizing a small fraction of parameters for downstream tasks, is receiving much attention \cite{houlsby2019parameter, lester2021power, li2021prefix, liu2022few, liu2023gpt}. Among various PEFT strategies, adapter-tuning has emerged as a prevalent method. It integrates lightweight modules, termed adapters, into each layer of PLMs and only tunes the adapters with the downstream tasks. As shown in Figure~\ref{fig_1}(a), the adapter-tuning methods \cite{houlsby2019parameter, pfeiffer2021adapterfusion, zaken2022bitfit, hu2021lora, zhang2022adaptive}, significantly reduce the number of trainable parameters compared to the full-tuning while exhibiting better performance on a downstream task.

As adapter-tuning reduces the number of trainable parameters, it is also expected to reduce the resource (i.e., memory) usage accordingly. Unfortunately, parameter-efficiency does not always translate into resource-efficiency. As shown in Figure~\ref{fig_1}(b), although adapter-tuning significantly reduces the number of trainable  parameters (by 99.37\%, on average) compared to the full-tuning, the memory usage is not much reduced (only by 22.19\%, on average). This is because adapter-tuning does not reduce activation memory (i.e., intermediate values for reuse during backpropagation) which account for 76.00\% of memory usage --- it only reduces optimizer memory (e.g., gradients and momentum vectors). Considering the remarkable increase in model size compared to the modest increase in GPU memory capacity, adapter-tuning methods still face challenges in terms of memory efficiency. For example, fine-tuning of a LLaMA-65B \cite{touvron2023llama} requires more than 780GB of GPU memory. As shown in Figure \ref{fig_1}(b), enabling resource-efficient fine-tuning may enhance accessibility of fine-tuning to researchers and end-users, by reducing memory requirements below the capacity of commodity GPU memory.

\begin{figure*}[t]
% \vskip -0.1in
\centering
\centerline{\includegraphics[width=\textwidth]{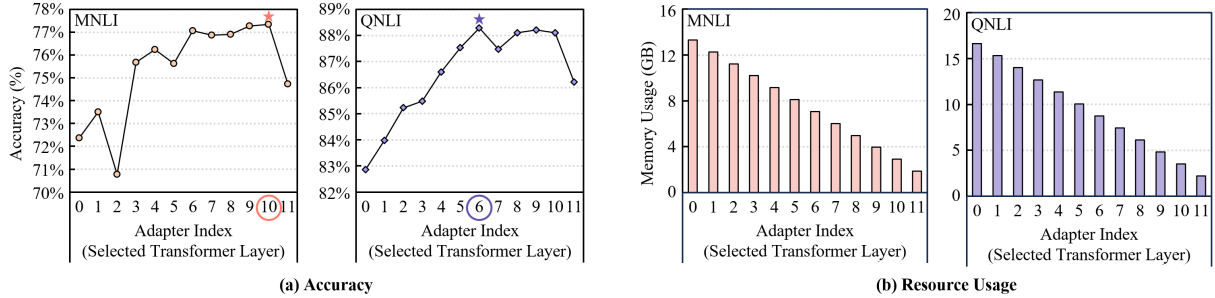}}
\caption{(a) Accuracy and (b) Resource usage of adapter-tuning by injecting an adapter into each transformer layer of $\text{BERT}_\text{base}$ model on MNLI and QNLI datasets from GLUE.}
\label{fig_2}
% \end{center}
\vskip -0.15in
\end{figure*}

According to previous work, the activation memory mostly depends on the backpropagation length \cite{chen2016training, rhu2016vdnn}, which is determined by the number of adapters trained during the backward pass. Hence, to reduce the activation memory, it is crucial to reduce the number of training adapters. However, merely reducing the number of training adapters degrades accuracy. Here, a pivotal research problem arises:
\begin{center}
\textbf{\textit{Can we reduce the number of training adapters without sacrificing accuracy?}}
\end{center}
To answer the question, we analyze the impact of individual adapters on the accuracy and resource usage of training (Figure~\ref{fig_2} in Section~\ref{Section 3}). We observe that some adapters are being trained, even after they finish contributing to the accuracy improvement, occupying memory. Thus, it is possible to stop training (i.e., freezing) such adapters early if they do not contribute to the adaptation for a downstream task, de-allocating their activation memory. We also observe that such early freezing can even lead to the regularization effect on the model~\cite{fu2023effectiveness}, improving the accuracy.

In this paper, we propose \textbf{SAFE} (\textbf{S}elective \textbf{A}dapter \textbf{F}r\textbf{E}ezing), which adaptively freezes adapters in the early epochs of training. In each epoch, SAFE identifies adapters that contribute relatively less to the accuracy improvement by using an importance score~\cite{kornblith2019similarity}. It then freezes the adapters whose importance score is lower than a pre-defined threshold, reducing the memory usage and accelerating training time. By early freezing less important adapters, SAFE induces regularization effect on the model being trained, leading to a flatter loss surface. This is beneficial for finding an optimal point with higher generalization performance while optimizing neural network. In our evaluation, SAFE significantly reduces the average memory usage and TFLOPs by 46.89\% and 51.73\%, respectively, across various models and downstream tasks without compromising accuracy compared to the baseline, LoRA~\cite{hu2021lora}. SAFE even improves the accuracy for some tasks, compared to LoRA, by up to 4.33\% while reducing memory usage by 53.60\%, by inducing the regularization effect.

In summary, our key contributions include:
\begin{compactitem}
    \item We uncover that adapters exhibit varying degrees of contribution to model adaptation and resource usage (Section \ref{Section 3}). 
    \item Motivated by this observation, we propose SAFE, a novel approach that enables resource-efficient fine-tuning by selectively early-freezing less important adapters (Section \ref{Section 4}).
    \item Our evaluation on various downstream tasks demonstrates that SAFE not only achieves comparable or even better task performance to baselines but also significantly reduces resource usage by inducing the regularization effect on the model (Section \ref{Section 5}).
\end{compactitem}

%%%%%%%%%%%%%%%%%%%%%%%%%%%%%%%%%%%%%%%%%%%%%%%%%%%%%%%%%
% RELATED WORK
%%%%%%%%%%%%%%%%%%%%%%%%%%%%%%%%%%%%%%%%%%%%%%%%%%%%%%%%%
\section{Related Work} \label{Section 2}
\textbf{Parameter-Efficient Fine-Tuning: } To efficiently adapt large-scale PLMs to downstream tasks, many adapter-tuning methods \cite{chen2023longlora, he2023parameter, hu2021lora, houlsby2019parameter, karimi2021compacter, liu2022few} have been proposed. In general, adapter-tuning methods inject small, trainable, and task-specific adapter modules into each transformer layer of a pre-trained model. Given a pre-trained weight matrix \( W_0 \in \mathbb{R}^{d \times k} \) and input \(x \in \mathbb{R}^{k \times 1}\), the weight update of adapter-tuning is expressed as \( W_0 + \Delta W \). During training, \( W_0 \) is frozen and does not receive gradient updates, while \( \Delta W \) contains trainable parameters. For \( h = W_0x \), The modified forward pass in adapter-tuning yields: 

\vspace{-10pt}
\begin{equation} \label{equation_1}
\setlength{\abovedisplayskip}{5pt}
\setlength{\belowdisplayskip}{-1pt}
h = W_0x + \Delta Wx.
\end{equation}
\vspace{-15pt}

To further improve parameter efficiency of adapter-tuning, AdaLoRA \cite{zhang2022adaptive} adaptively adjusts the number of trainable parameters among adapters according to their importance score --- it reduces the number of trainable parameters for less important adapters. However, the adapter-tuning methods still use a large amount of memory, as shown in Figure~\ref{fig_1}, since they do not reduce the activation memory which accounts for a large portion of memory usage.

\noindent
\textbf{Pruning LLM Model Parameters: } To reduce the model memory of fine-tuning, two categories of pruning methods have been proposed~\cite{liang2021pruning}: structured pruning and unstructured pruning. Structured pruning methods remove grouped parameters (e.g., channels, layers) from the LLM. However, they usually degrade the accuracy. Furthermore, they have a limitation in terms of the compression ratio because of the low flexibility. LLM-Pruner \cite{ma2023llm} compensates the accuracy drop coming from pruning, by employing post-training. 

To overcome the limitation of structured pruning, unstructured pruning methods remove partial values of weight matrices regardless of their structures \cite{li2022parameter, frantar2023sparsegpt}. However, unstructured pruning also degrades the accuracy.

%can obtain highly compressed models by directly pruning individual parameters. However, unstructured pruning requires the support of hardware architecture with specific design.
% However, resource efficiency of unstructured pruning relies on sparse operations hardware support. 

\noindent
\textbf{Resource Efficient Fine-Tuning: } Several works have tried to target resource efficient fine-tuning. AdapterDrop \cite{ruckle2021adapterdrop}, randomly excludes partial adapters from each training step. However, it cannot de-allocate the activation memory for the adapters, because of the random selections --- an adapter excluded from training in a step can be included in training in the following steps. SparseAdapter\cite{he2022sparseadapter} applies unstructured pruning to the adapters. However, it also does not reduce the actual memory usage --- this is because the weight matrices pruned with zero values still need to be fully allocated in the memory. LoRAPrune\cite{zhang2024loraprune} employs structured pruning for LoRA. Unfortunately, the aforementioned methods usually have an adverse impact on the accuracy.
MEFT \cite{liao2024make} applies a reversible model to PEFT. By using the reversible network, MEFT calculates activations with accumulated outputs of layers, without saving the intermediate activations reducing the activation memory. However, calculations of the activations severely degrades the training time performance. 

Different from the previous works, this work freezes less important adapters in early steps of training. Since the frozen adapters can only be used for the forward pass, early freezing of less important adapters can effectively reduce the backpropagation length as well as the activation memory. Moreover, it induces regularization effect on the model, improving its accuracy.

%%%%%%%%%%%%%%%%%%%%%%%%%%%%%%%%%%%%%%%%%%%%%%%%%%%%%%%%%
% Background and Motivation
%%%%%%%%%%%%%%%%%%%%%%%%%%%%%%%%%%%%%%%%%%%%%%%%%%%%%%%%%
\section{Motivation} \label{Section 3}
In this section, we present a pivotal research question for resource-efficient fine-tuning. 

\begin{rqbox}
 \textbf{RQ}: Do all adapters contribute equally to the process of adaptation?
\end{rqbox}
To answer this question, we analyze the impact of adapters injected into each transformer layer on accuracy and resource efficiency. We measure the accuracy and memory usage of $\text{BERT}_\text{base}$ model on MNLI and QNLI dataset from GLUE~\cite{wang2018glue}, by attaching an adapter to each transformer layer one-by-one. Figure \ref{fig_2}(a) and (b) show the measured accuracy and memory usage respectively --- the x-axis indicates the index of transformer layer that the adapter is injected into.

As shown in Figure~\ref{fig_2}(a), each adapter has different impact on the accuracy, and the importance of each adapter varies depending on the downstream task. In addition, despite uniform counts of trainable parameters, resource usage decreases for adapters closer to the output layer, as depicted in Figure~\ref{fig_2}(b). These observations point to the possibility that adapters in early layers contribute less to task adaptation, even though they require considerable resources. In other words, if we selectively deactivate less impactful adapters, it is possible to co-optimize the resource efficiency and accuracy.

\begin{figure}[t]
% \vskip -0.1in
\centering
\centerline{\includegraphics[width=\columnwidth]{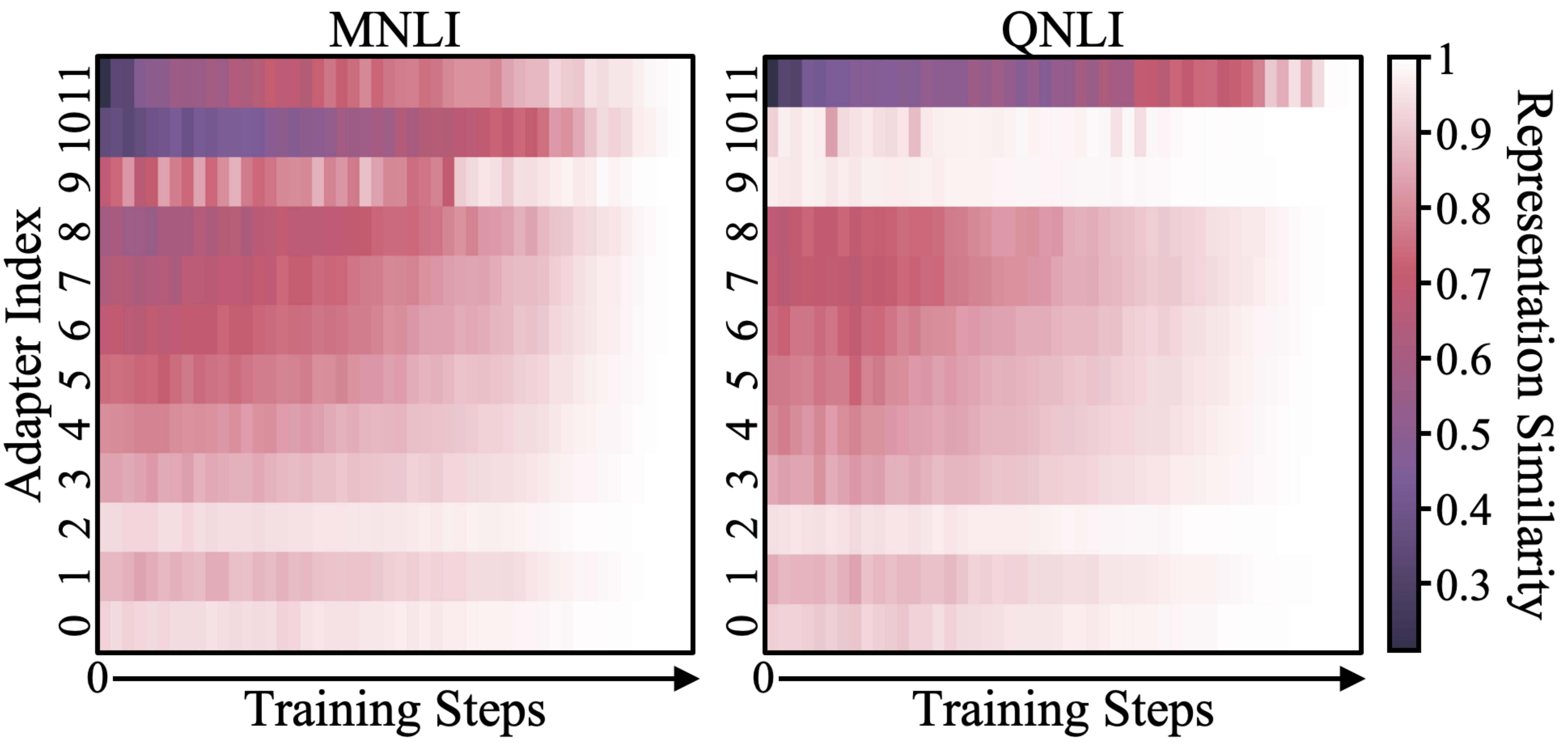}}
\caption{Visualization of representation similarity between the trained model and the model at different training steps during adapter-tuning of the $\text{BERT}_\text{base}$ model on the MNLI and QNLI datasets from GLUE.}
\label{fig_3}
% \end{center}
\vskip -0.1in
\end{figure}

To further analyze changes of the feature representations for each adapter throughout the training process, we quantify the representation similarity between adapters in each training step and those in the final model (which we obtained after the convergence of fine-tuning). We quantify the representational similarity using Centered Kernel Alignment (CKA) by referring to previous works~\cite{li2022smartfrz}. Figure~\ref{fig_3} visualizes the representational similarity measured throughout the training process for each adapter for $\text{BERT}_\text{base}$ model on MNLI and QNLI dataset from GLUE~\cite{wang2018glue} --- lighter the color becomes, higher the feature representation similarity is.

As shown in Figure~\ref{fig_3}, even in the early training steps, the feature representations of several adapters are almost the same as those of the final model --- similar patterns are observed in other models and datasets. This means that those adapters are already representing the features that should be represented by the final model, and thus they may not further be adapted for the downstream task in the rest of the training steps. This is why such adapters are less contributing to the accuracy improvement of the model on the downstream tasks in Figure~\ref{fig_2}. One intuition is that lower adapters generally learn basic understanding of the input data, such as data bias and structural characteristics of the data, while adapters closer to the output build features unique to different tasks \cite{houlsby2019parameter}. Motivated by the observation where not all adapters consistently contribute to adaptation, in the next section, we propose a selective adapter freezing method which preemptively freezes adapters that are relatively less important for each task.
 
%%%%%%%%%%%%%%%%%%%%%%%%%%%%%%%%%%%%%%%%%%%%%%%%%%%%%%%%%
% Selective Adapter Freezing (SAFE)
%%%%%%%%%%%%%%%%%%%%%%%%%%%%%%%%%%%%%%%%%%%%%%%%%%%%%%%%%
\section{Selective Adapter Freezing (SAFE)} \label{Section 4}

In this section, we propose a selective adapter freezing method, SAFE. SAFE adaptively freezes less important adapters in the early training steps, in order to reduce unnecessary computation and memory usage without compromising the accuracy. 

\begin{figure*}[t]
% \vskip -0.1in
\begin{center}
\centerline{\includegraphics[width=\linewidth]{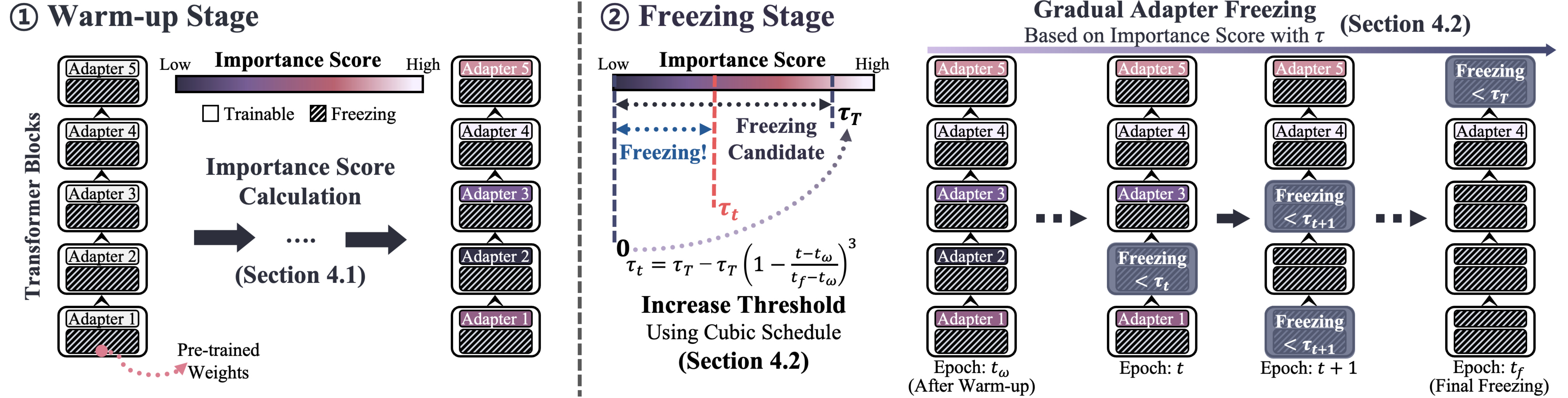}}
\caption{Design overview of Selective Adapter Freezing (SAFE). At the warm-up stage, SAFE identifies important adapters by calculating importance score. At the freezing stage, SAFE gradually freezes the adapter based on their importance score with moving threshold $\tau$ by following a cubic schedule.}
\label{fig_overview}
\end{center}
\vskip -0.0in
\end{figure*}

Figure~\ref{fig_overview} shows the overview of SAFE. SAFE consists of two stages: warm-up stage and freezing stage. In the warm-up stage, SAFE performs several epochs of fine-tuning while monitoring the feature representation changes (i.e., importance score) of the adapters (Section~\ref{Section 4.1}). If the important score of all adapters is not much changed for consecutive epochs, SAFE enters the freezing stage. In the freezing stage, SAFE gradually freezes adapters that contribute less to the adaptation, based on the importance score (Section~\ref{Section 4.2}). By early freezing less important adapters, SAFE induces regularization effect on model (Section~\ref{Section 4.3}), leading to better performance. 

\subsection{Importance Score} \label{Section 4.1}
In the warm-up stage\footnote{We define the number of warm-up epochs as the epoch at which the importance score of all adapters change by less than 5\% for consecutive epochs.}, we identify less important adapters by monitoring the feature representation changes of the adapters. To capture the feature representation changes of the adapters, SAFE uses Centered Kernel Alignment (CKA), which is a representative metric for representation similarity --- similar practice has been used in previous works~\cite{neyshabur2020being, raghu2021vision}. It calculates CKA between the activation of a layer adapted with an adapter and that of the original layer as:

\vskip -0.2in
\begin{align} \label{equation_5}
\text{CKA}_{i}(X_i, Y_i) = \frac{\|Y_i^T X_i\|_F^2}{\|X_i^T X_i\|_F \|Y_i^T Y_i\|_F},
% \vskip -0.2in
\end{align} 
where \(X_i\) and \(Y_i\) are the activations of a layer that is adapted with an adapter and the original layer, respectively, \(i = \text{Index of Layer}\), and \(\| \cdot \|_F^2\) represents the square of the Frobenius norm of a matrix.

Higher CKA value indicates that the feature representation of a layer is still similar with that of the original one. To this end, SAFE calculates the importance score of an adapter as:
\vskip -0.2in
\begin{align} \label{equation_6}
Imp(Adapter_{i}) = 1 - \text{CKA}_{i}(X_i, Y_i)
\end{align}

\subsection{Adapter Freezing} \label{Section 4.2}
In the freezing stage, SAFE gradually freezes adapters based on their importance score. At \(t_w\)-th epoch, SAFE compares the importance score of adapters with threshold $\tau_T$. If the importance score of an adapter is lower than $\tau_T$, SAFE identifies the adapter as a freezing candidate. After identifying freezing candidates, SAFE freezes them based on a moving threshold until \(t_f\)-th epoch --- it increases the threshold from 0 to $\tau_T$\footnote{We empirically determine $\tau_T$ and final freezing epochs $t_f$ based on extensive experiments with various models and datasets.} between \(t_w\)-th and \(t_f\)-th epochs following a cubic schedule as \cite{zhang2022adaptive}:
\begin{equation} \label{equation_7}
    \small
    \tau_t = 
    \begin{cases}
    0 & 0 \leq t < t_w, \\
    \tau_T - \tau_T \left(1 - \frac{t - t_w}{t_f - t_w}\right)^3 & t_w \leq t < t_f, \\
    \tau_T & \text{o.w.}
    \end{cases}
\end{equation}
where \(t\) is the current epoch, \(t_w\) is the number of initial warm-up epochs, \(t_f\) is the number of final freezing epochs. The cubic schedule follows a non-linear growth pattern, enabling rapid exploration of the search space during initial stages and gradually slowing down to reliably converge to the target point. We leverage these advantages by freezing more less-important adapters in the early stages. As the number of trainable parameters in the model decreases, SAFE gradually reduces the number of freezing adapters, reaching the threshold $\tau_T$ stably.
% This ensures that a large number of adapters participate in adaptation during the early stages of training, allowing SAFE to more accurately identify important adapters. As training progresses, less important adapters are gradually frozen, with different training periods assigned to each adapter based on their importance.
% assigning different training periods to each adapter based on their importance.

\subsection{Regularization Effect of SAFE} \label{Section 4.3}
By selectively freezing less critical adapters, SAFE induces a regularization effect within the model.
In transformer-based PLM \(\mathcal{N}_0\), each of the \(l\) transformer blocks \(T_l\) is equipped with a distinct set of parameters \(\theta^0_l\) for \(l \in \{1, \ldots, n\}\). 
To reduce the computational overhead of fine-tuning all parameters \(\theta^0_l\), lightweight adapters \(\Delta \theta_l\) are introduced.
% The introduction of adapters not only reduces training time but also enhances performance on downstream tasks.
To clarify how introducing adapters contribute to performance enhancements, \citet{fu2023effectiveness} formalize the optimization function as follows:
\begin{equation} \label{eq:regularization}
    \min_{\theta} \mathcal{L}(\theta) + \| (I - M)(\theta - \theta^0) \|^2,
\end{equation}
where \(\theta = \theta^0 + M\Delta \theta\) and \(M \in \{0,1\}^{m \times m}\), with \(m=\text{dim}(\theta)\), serves as a diagonal matrix for selective parameter adjustment. 
Each diagonal element \(M_{ii} \in \{0,1\}\) indicates whether the corresponding parameter of \(\Delta \theta_i\) is active (\(1\)) or inactive (\(0\)), with all off-diagonal elements \(M_{ij}\) set to 0.
The regularization term is crucial for explaining how parameter constraints introduced by adapters can enhance model performance on downstream tasks.
The $rank(M)$ is bounded by \(m\), reflecting full capacity for parameter adaptation within each transformer block.
However, such an approach can lead to excessive computation (See Figure~\ref{fig_1}). 
% In contrast, our study explores the implications of constraining the $rank(M)$ to a reduced upper limit of \(\frac{m}{l}\) by selectively activating $\Delta W_l$ for $T_l$.
% This constraint not only optimizes computational efficiency but also preserves the adaptability essential for superior performance on downstream tasks, as evidenced by our empirical results detailed in Section~\ref{Section 5.3}.
In contrast, our study explores the implications of constraining the $rank(M)$ to a reduced upper bound from $m$ to the number of trainable parameters following the proposed freezing algorithm in Section~\ref{Section 4.2}. For instance, in our motivational analysis in Section~\ref{Section 3}, where we limit the number of trainable parameters to one adapter per layer, the $rank(M)$ is bounded by $\frac{m}{l}$ by selectively activating $\Delta W_l$ for $T_l$. This constraint not only optimizes computational efficiency but also preserves the adaptability essential for superior performance on downstream tasks, as evidenced by our empirical results detailed in Section~\ref{Section 5.3}.

%%%%%%%%%%%%%%%%%%%%%%%%%%%%%%%%%%%%%%%%%%%%%%%%%%%%%%%%%
% Experiments
%%%%%%%%%%%%%%%%%%%%%%%%%%%%%%%%%%%%%%%%%%%%%%%%%%%%%%%%%

%%%%%%%%%%%%%%%%%%%%%%%%%%%%%%%%%%%%%%%%%%%%%%%%%%%%%%%%%%
% TABLE_BEGIN
%%%%%%%%%%%%%%%%%%%%%%%%%%%%%%%%%%%%%%%%%%%%%%%%%%%%%%%%%%
\begin{table*}[t]
\centering
\renewcommand{\arraystretch}{1.4}
\caption{Experimental results with \(\text{BERT}_\text{large}\) on natural language understanding tasks from the GLUE benchmark. SAFE significantly reduces memory usage while achieving GLUE score comparable to the baseline. Note that we report memory usage and computation costs on RTE task.}
\label{table_1}
% \footnotesize
\fontsize{8}{9}\selectfont 
% \tiny
\setlength\tabcolsep{1pt}
\begin{tabular}{@{}l|cccccccc|ccc@{}}
\toprule
     &
      CoLA &
      SST-2 &
      MNLI &
      RTE &
      QQP &
      MRPC &
      QNLI &
      STS-B &
      Avg. &
      Memory &
      Computation \\
 &
  Matthews corr &
  Accuracy &
  Accuracy &
  Accuracy &
  Accuracy &
  F1 Score &
  Accuracy &
  Pearson   corr &
   &
   (GB) &
   (TFLOPs) \\ \bottomrule
LoRA &
  65.24 &
  93.65 &
  85.40 &
  72.66 &
  90.49 &
  87.90 &
  90.06 &
  91.88 &
  84.66 &
  20.35 &
  46,698 \\
+   \citet{zhangrevisiting} &
  61.78 &
  91.97 &
  84.61 &
  73.18 &
  89.20 &
  86.83 &
  90.00 &
  87.50 &
  83.13 &
  11.20 &
  35,415 \\
+   AdapterDrop &
  64.24 &
  92.54 &
  85.19 &
  73.38 &
  89.02 &
  86.51 &
  91.51 &
  91.39 &
  84.22 &
  20.35 &
  35,114 \\
+   SparseAdapter &
  65.25 &
  92.66 &
  85.19 &
  74.10 &
  90.43 &
  88.50 &
  91.61 &
  91.99 &
  84.97 &
  20.35 &
  46,698 \\
+   LoRAPrune &
  63.03 &
  91.54 &
  83.97 &
  70.59 &
  88.28 &
  87.74 &
  84.13 &
  86.48 &
  81.97 &
  10.37 &
  25,137 \\ 
+   MEFT &
  64.57 &
  92.66 &
  84.33 &
  72.92 &
  88.70 &
  88.30 &
  90.98 &
  90.15 &
  84.08 &
  11.15 &
  88,363 \\ \hline
    \cellcolor[HTML]{F2F2F2}+ SAFE &
  \cellcolor[HTML]{F2F2F2}65.26 &
  \cellcolor[HTML]{F2F2F2}92.78 &
  \cellcolor[HTML]{F2F2F2}85.41 &
  \cellcolor[HTML]{F2F2F2}74.10 &
  \cellcolor[HTML]{F2F2F2}89.96 &
  \cellcolor[HTML]{F2F2F2}88.84 &
  \cellcolor[HTML]{F2F2F2}91.78 &
  \cellcolor[HTML]{F2F2F2}91.80 &
  \cellcolor[HTML]{F2F2F2}\textbf{84.99} &
  \cellcolor[HTML]{F2F2F2} \textbf{12.11{\fontsize{5pt}{6pt}\selectfont\textcolor{blue}{40.47\%$\downarrow$}}}&
  \cellcolor[HTML]{F2F2F2} \textbf{30,285{\fontsize{5pt}{6pt}\selectfont\textcolor{blue}{35.15\%$\downarrow$}}}\\ \bottomrule
\end{tabular}
\end{table*}

%%%%%%%%%%%%%%%%%%%%%%%%%%%%%%%%%%%%%%%%%%%%%%%%%%%%%%%%%%
% TABLE_END 
%%%%%%%%%%%%%%%%%%%%%%%%%%%%%%%%%%%%%%%%%%%%%%%%%%%%%%%%%%

\section{Experiments} \label{Section 5}
\subsection{Experimental Setting} \label{Section 5.1}
\textbf{Models:} We assess the fine-tuning efficacy of SAFE using state-of-the-art transformer-based models, including \(\text{BERT}_\text{base}\), \(\text{BERT}_\text{large}\)~\cite{kenton2019bert}, 
\(\text{RoBERTa}_\text{base}\), \(\text{RoBERTa}_\text{large}\)~\cite{liu2019roberta}, \(\text{GPT-2}_\text{medium}\), \(\text{GPT-2}_\text{large}\)~\cite{radford2019language}, and \(\text{LLaMA-2}_\text{7B}\)~\cite{touvron2023llama}. 

\noindent
\textbf{Datasets:} The aforementioned models are evaluated across various tasks that span a broad spectrum of NLP applications, including Natural Language Understanding (NLU), Question Answering (QA), and Natural Language Generation (NLG). Initially, we utilize eight datasets from the General Language Understanding Evaluation (GLUE)~\cite{wang2018glue} which comprises two single-sentence classification tasks, three similarity and paraphrase tasks, and four natural language inference tasks. Furthermore, we conduct experiments on the SQuAD dataset~\cite{rajpurkar2016squad} with both BERT and RoBERTa model families. Decoder-only models such as \(\text{GPT-2}_\text{large}\) are also tested to determine if SAFE maintains its effectiveness in the E2E NLG Challenge~\cite{novikova2017e2e}. Finally, to investigate the scalability to larger models, we evaluate SAFE on the large language model (\(\text{LLaMA-2}_\text{7B}\)) and the WikiText-2 dataset~\cite{merity2022pointer}. Detailed dataset descriptions are available in Appendix \ref{Dataset Statistics}.

\noindent
\textbf{Baselines:} To evaluate the effectiveness of SAFE, we benchmark against state-of-the-art PEFT method, LoRA~\cite{hu2021lora}. We compare SAFE with five effective resource efficient fine-tuning methods, a previous work~\cite{zhangrevisiting}, AdapterDrop~\cite{ruckle2021adapterdrop}, SparseAdapter~\cite{he2022sparseadapter}, LoRAPrune~\cite{zhang2024loraprune}, and MEFT~\cite{liao2024make}. SAFE's performance is further compared with four other PEFT methods such as Houlsby \cite{houlsby2019parameter}, Pfeiffer \cite{pfeiffer2020adapterhub}, BitFit \cite{zaken2022bitfit}, and the adaptive method AdaLoRA \cite{zhang2022adaptive} to demonstrate its versatility and applicability across different adapter-tuning frameworks (see Figure \ref{fig_1} and Appendix \ref{Results with Various PEFT Methods} for detailed results). Comprehensive details on the experimental setup and hyperparameters, such as training epochs and batch sizes, can be found in Appendix \ref{Experimental Setups}.

\subsection{Main results} \label{Section 5.2} 
\subsubsection{Natural Language Understanding} 
Table \ref{table_1} shows the results of different methods on GLUE tasks. Since SAFE selectively freezes 51.04\% of less important adapters early throughout the training process, SAFE significantly reduces memory usage by 40.47\%, from 20.35GB (LoRA) to 12.11GB, and decreases computation costs (FLOPs) by 35.15\%. Even with such improvements in resource efficiency, SAFE improves the average GLUE score from 84.66 (LoRA) to 84.99 --- this is because SAFE induces a regularization effect on less-important adapters improving generalization performance of the model (see Section~\ref{Section 5.3}).

\begin{figure}[t]
% \vskip -0.1in
\centering
\centerline{\includegraphics[width=\columnwidth]{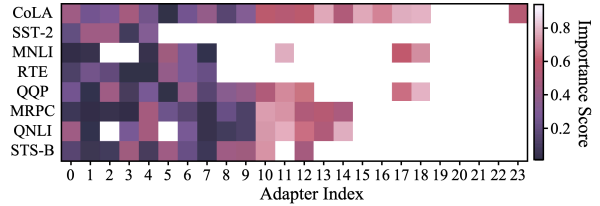}}
\caption{The freezing patterns when fine-tuning $\text{BERT}_\text{large}$ on GLUE with SAFE. Colors indicate adapters that are frozen, while white represents an adapter that is not frozen --- the lighter the color is, the higher importance score is.}
\label{fig_5}
% \end{center}
% \vskip -0.1in
\end{figure}

Figure \ref{fig_5} shows the freezing patterns of $\text{BERT}_\text{large}$ fine-tuned with SAFE --- we observe similar patterns for other tasks. We find that SAFE tends to freeze adapters more in layers closer to the input layer. Such behavior aligns with empirical observations presented in Figure \ref{fig_2} where adapters closer to the output layer need further adaptation to the downstream tasks compared to those in earlier layers, contributing more to model performance.

Compared to a previous work~\cite{zhangrevisiting}, which attaches adapters only to the upper layers, SAFE achieves 2.24\% higher GLUE score. This is because some adapters attached to lower layers keep contributing to the adaptation even in later training epochs, as shown in Figure~\ref{fig_5} (e.g., 2nd and 5th adapters in QNLI task). Compared to AdapterDrop, SAFE provides up to 2.69\% higher score (MRPC) while reducing memory usage by 49.47\% (0.91\% higher GLUE score and 40.47\% reduced memory usage on average). This is because AdapterDrop reduces computation costs (FLOPs) by randomly dropping adapters for each step, whereas SAFE selectively freezes less-important adapters preserving critical adapters and thus achieving better performance. AdapterDrop also does not lower memory usage because the memory allocated to the dropped adapters cannot be de-allocated for the next step --- adapters dropped in a step may not be dropped in the next step. Compared to SparseAdapter, SAFE reduces memory usage by 40.47\% and computation cost by 35.15\% while providing comparable GLUE score. This is because SparseAdapter performs pruning of redundant parameters but uses unstructured pruning with masking, which does not actually improve resource efficiency. On the other hand, although LoRAPrune is effective to reduce the memory usage, it severely degrades the accuracy as it fully eliminates less important adapter weights through structural pruning. As a result, SAFE achieves 3.68\% higher GLUE score compared to LoRAPrune. MEFT reduces the memory usage at the cost of more than 2x of the FLOPs and training time. This is because MEFT applies a reversible model to LoRA, which reduces memory usage by not caching intermediate activations, but causes substantial computation overhead due to recomputation. Overall, SAFE strikes a better accuracy/memory efficiency/training-time performance trade-off compared to SOTA methods.

%Compared to LoRAPrune, SAFE achieves a 3.68\% higher GLUE score. This is because LoRAPrune fully eliminates less important adapter weights through structural pruning, leading to significant accuracy degradation. 

\begin{figure}[t]
\begin{center}
\centerline{\includegraphics[width=\columnwidth]{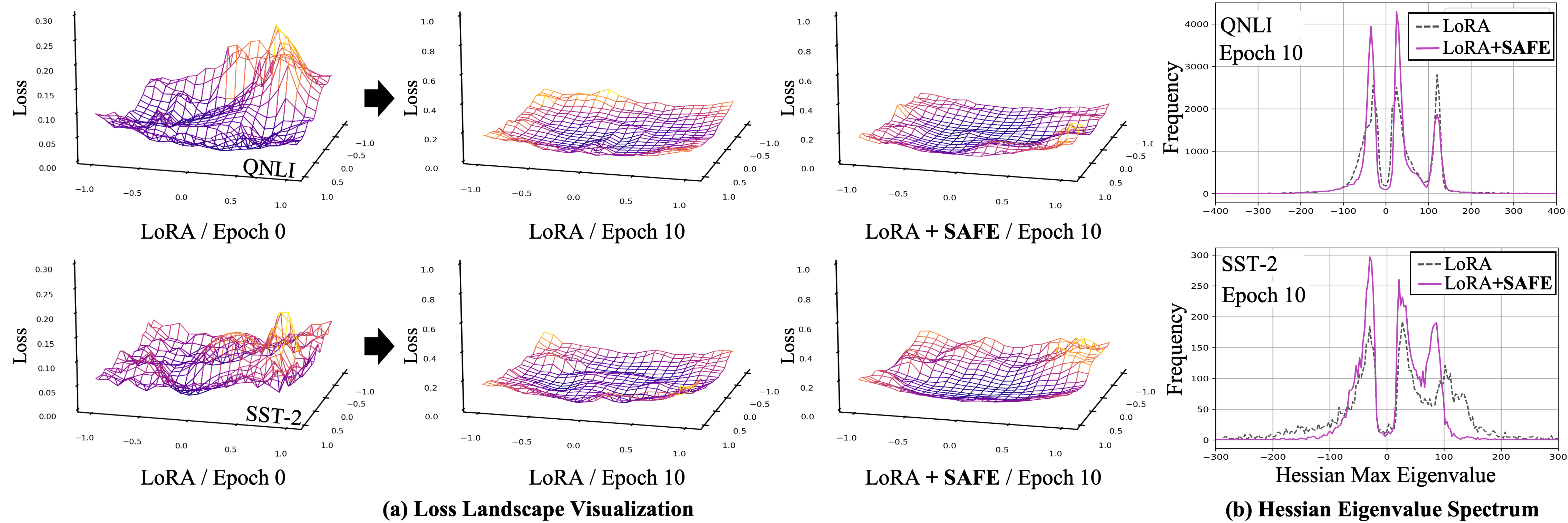}}
\caption{Comparison of (a) perplexity and (b) resource usage between LoRA and SAFE on the $\text{LLaMA-2}_\text{7B}$ model evaluated on the WikiText-2 dataset.}
\label{fig_6}
\end{center}
\vskip -0.4in
\end{figure}

\begin{table}[t]
    \centering
    \captionof{table}{Experimental results on question answering task from the SQuAD dataset.}
    \label{table_2}
    % \scriptsize
    \fontsize{8}{9}\selectfont 
    \setlength\tabcolsep{2pt}
    \renewcommand{\arraystretch}{1.1}
\begin{tabular}{cl|c|cc}
\toprule
 &
  \multicolumn{1}{c|}{} &
  F1 Score &
  \begin{tabular}[c]{@{}c@{}}Memory Usage\\      (GB)\end{tabular} &
  \begin{tabular}[c]{@{}c@{}}Computation\\      (TFLOPs)\end{tabular} \\ \bottomrule
\multirow{2}{*}{$\text{BERT}_\text{base}$}     
    & LoRA   
    & 86.99          
    & 5.95
    & 611,295 \\
    
    & \cellcolor[HTML]{F2F2F2}+ SAFE 
    & \cellcolor[HTML]{F2F2F2}\textbf{87.22} 
    & \textbf{\cellcolor[HTML]{F2F2F2}4.61{\fontsize{5pt}{6pt}\selectfont\textcolor{blue}{22.52\%$\downarrow$}}   } 
    & \textbf{\cellcolor[HTML]{F2F2F2}455,274{\fontsize{5pt}{6pt}\selectfont\textcolor{blue}{25.52\%$\downarrow$}}} \\ \midrule
\multirow{2}{*}{$\text{BERT}_\text{large}$}    
    & LoRA   
    & 89.22          
    & 15.79 
    & 2,117,791 \\
                               
    & \cellcolor[HTML]{F2F2F2}+ SAFE 
    & \cellcolor[HTML]{F2F2F2}\textbf{89.72} 
    & \textbf{\cellcolor[HTML]{F2F2F2}7.44{\fontsize{5pt}{6pt}\selectfont\textcolor{blue}{52.88\%$\downarrow$}}}    
    & \textbf{\cellcolor[HTML]{F2F2F2}869,624{\fontsize{5pt}{6pt}\selectfont\textcolor{blue}{58.94\%$\downarrow$}}} \\ \midrule
\multirow{2}{*}{$\text{RoBERTa}_\text{base}$}  
    & LoRA   
    & 90.95          
    & 11.51 
    & 1,225,147 \\

    & \cellcolor[HTML]{F2F2F2}+ SAFE 
    & \cellcolor[HTML]{F2F2F2}\textbf{91.16} 
    & \textbf{\cellcolor[HTML]{F2F2F2}7.80{\fontsize{5pt}{6pt}\selectfont\textcolor{blue}{32.23\%$\downarrow$}}}  
    & \textbf{\cellcolor[HTML]{F2F2F2}808,239{\fontsize{5pt}{6pt}\selectfont\textcolor{blue}{34.03\%$\downarrow$}}} \\ \midrule
\multirow{2}{*}{$\text{RoBERTa}_\text{large}$} 
    & LoRA   
    & 93.39          
    & 17.73 
    & 2,117,791 \\ 
    
    & \cellcolor[HTML]{F2F2F2}+ SAFE 
    & \cellcolor[HTML]{F2F2F2}\textbf{94.13} 
    & \textbf{\cellcolor[HTML]{F2F2F2}3.56{\fontsize{5pt}{6pt}\selectfont\textcolor{blue}{79.92\%$\downarrow$}}}  
    & \textbf{\cellcolor[HTML]{F2F2F2}245,541{\fontsize{5pt}{6pt}\selectfont\textcolor{blue}{88.41\%$\downarrow$}}} \\ \bottomrule
\end{tabular}
\end{table}

% \noindent
\subsubsection{Question Answering} 
Table \ref{table_2} shows the results of SQuAD dataset. SAFE consistently outperforms baseline under all settings. Notably, SAFE reduces memory usage and computation costs by up to 79.92\% and 88.41\% on $\text{RoBERTa}_\text{large}$ by freezing 91.67\% of the adapters, while improving the F1 score from 93.39 (LoRA) to 94.13. This result also demonstrates that the benefits and effectiveness of SAFE are not restricted to specific model sizes, making it a valuable strategy for enhancing adapter-tuning outcomes across models of varying scales.

% \noindent

\begin{table}[t]
    \centering
    % \renewcommand{\arraystretch}{1.1}
    % \vskip -0.17in
    \captionof{table}{Experimental results on natural language generation from the E2E NLG Challenge. For all metrics, higher is better. Note that we report memory usage reduction in \textcolor{blue}{blue}.}
    % \vskip 0.3in
    \label{table_3}
    \fontsize{7}{10}\selectfont 
    % \scriptsize
    % \tiny
    \setlength\tabcolsep{2pt}
    \begin{tabular}{@{}cl|ccccc|c@{}}
    \toprule
    \multicolumn{1}{l}{} &
       &
      \multicolumn{1}{c}{BLEU} &
      \multicolumn{1}{c}{NIST} &
      \multicolumn{1}{c}{METEOR} &
      \multicolumn{1}{c}{ROUGE-L} &
      \multicolumn{1}{c}{CIDEr} &
      \multicolumn{1}{|c}{} \\   \bottomrule
    \multirow{2}{*}{$\text{GPT-2}_{\text{medium}}$} 
        & LoRA 
        &  68.91
        &  8.68
        &  46.48
        &  71.33
        &  2.47
        &  \\
        & \cellcolor[HTML]{F2F2F2}+ SAFE     
        & \cellcolor[HTML]{F2F2F2}68.67
        & \cellcolor[HTML]{F2F2F2}8.66
        & \cellcolor[HTML]{F2F2F2}46.40 
        & \cellcolor[HTML]{F2F2F2}70.88 
        & \cellcolor[HTML]{F2F2F2}2.43 
        & \cellcolor[HTML]{F2F2F2}{\fontsize{5pt}{6pt}\selectfont{\textcolor{blue}{34.34\%$\downarrow$}}}
    \\ \midrule
    \multirow{2}{*}{$\text{GPT-2}_{\text{large}}$} 
    & LoRA
        &  70.27
        &  8.85
        &  46.40
        &  71.63
        &  2.52 
        &  \\
    & \cellcolor[HTML]{F2F2F2}+ SAFE     
        & \cellcolor[HTML]{F2F2F2}70.26
        & \cellcolor[HTML]{F2F2F2}8.87
        & \cellcolor[HTML]{F2F2F2}46.58
        & \cellcolor[HTML]{F2F2F2}71.68 
        & \cellcolor[HTML]{F2F2F2}2.53
        & \cellcolor[HTML]{F2F2F2}{\fontsize{5pt}{6pt}\selectfont{\textcolor{blue}{25.50\%$\downarrow$}}}
    \\ \bottomrule 
    \end{tabular}
\end{table}

\FloatBarrier

\begin{figure*}[t]
    % \vskip -0.2in
    \begin{center}
    \centerline{\includegraphics[width=\linewidth]{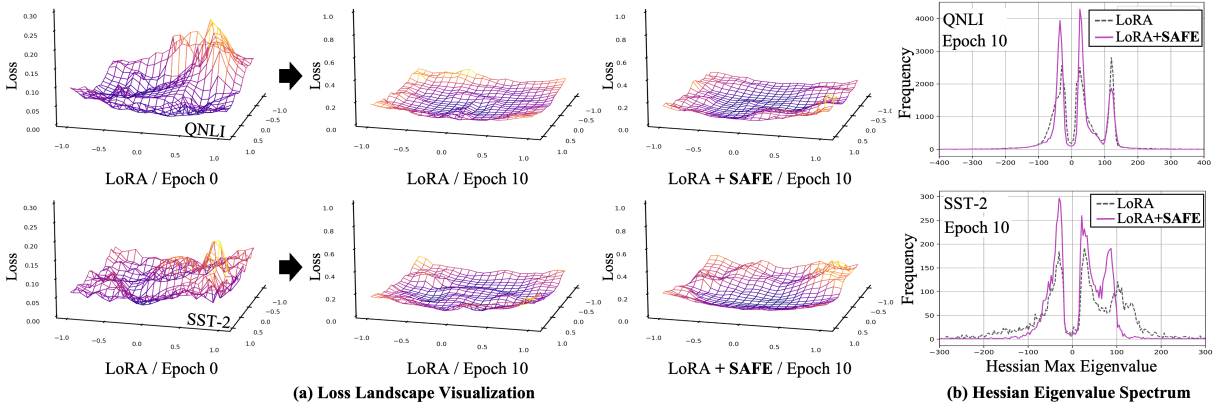}}
    \caption{
    %(a) Loss landscape visualization. SAFE has a flatter loss surface. 
    (a) Loss landscape demonstrates that SAFE yields a flatter loss surface compared to the baseline, as shown in the second and third columns. 
    %(b) Hessian eigenvalues spectrum. The magnitude of the Hessian eigenvalues of SAFE is smaller than the baseline. Since the Hessian represents local curvature, this also suggests that the loss landscapes of SAFE is flatter than the baseline.
    (b) Hessian eigenvalue spectrum analysis shows that the magnitude of the Hessian eigenvalues for SAFE is smaller than those for the baseline, indicating a flatter local curvature and potentially better generalization properties.
    }
    \label{fig_7}
    \end{center}
    % \vskip -0.2in
\end{figure*}

\subsubsection{Natural Language Generation}
Table \ref{table_3} shows that SAFE prevails on natural language generation task with $\text{GPT-2}$. SAFE achieves comparable performance to LoRA across all metrics while significantly reducing memory usage. This result also demonstrates that SAFE is effective not only for encoder models but also works well with decoder models.

% To examine the applicability of SAFE on state-of-the-art large language models, we compare SAFE with LoRA for a large language model, $\text{LLaMA-2}_\text{7B}$, using the WikiText-2 dataset. Figure~\ref{fig_6} shows that SAFE reduces memory usage by 48.37\%, from 66.35GB (LoRA) to 34.20GB, without any degradation in model quality. This result implies that SAFE can significantly improve the scalability of large language models by reducing the memory footprint, making it feasible to train these models on resource-constrained devices or with larger batch sizes.

% Figure~\ref{fig_6} shows the comparison of SAFE with LoRA for a large language model, $\text{LLaMA-2}_\text{7B}$, using the WikiText-2 dataset to examine the applicability of SAFE on state-of-the-art large language models. As shown in Figure~\ref{fig_6}, SAFE reduces memory usage by 48.37\%, from 66.35GB (LoRA) to 34.20GB, without any degradation in model quality. This result implies that SAFE can significantly improve the scalability of large language models by reducing the resource usage, making it feasible to train these models on resource-constrained devices or with larger batch sizes.

Figure~\ref{fig_6} compares (a) perplexity and (b) memory usage between LoRA and SAFE on the $\text{LLaMA-2}_\text{7B}$ using the WikiText-2 dataset, to examine the applicability of SAFE on large language models. As shown in Figure~\ref{fig_6}, SAFE reduces memory usage by 48.37\%, from 66.35GB (LoRA) to 34.20GB, without any degradation in model quality. This result implies that SAFE can significantly improve the scalability of large language models by reducing the resource usage, making it feasible to train these models on resource-constrained devices or with larger batch sizes.

\subsection{Regularization Effect} \label{Section 5.3}
To elucidate the underlying mechanisms behind SAFE's enhancements in model performance and memory efficiency, we conduct a detailed empirical analysis. We visualize and compare the loss landscapes of the baseline and SAFE. Additionally, we quantitatively evaluate the flatness of the loss surfaces by analyzing the spectrum of Hessian eigenvalues. This methodical approach allows us to substantiate the improvements attributed to SAFE, providing insights into its effectiveness in optimizing both performance and resource utilization.

\noindent
\textbf{Loss Landscape Analysis.} The flatness of a loss landscape is a recognized indicator of the generalization ability of models \cite{jiangfantastic}. Specifically, flatter landscapes are indicative of enhanced robustness to parameter perturbations \cite{xie2021artificial}, reduced model complexity \cite{blier2018description}, and improved generalization capabilities \cite{cha2021swad, cha2022domain, choromanska2015loss, park2021vision, wu2023implicit}. To examine these properties, we employ a comparative visualization of the loss landscapes for LoRA and SAFE using the $\text{BERT}_\text{base}$ model on the QNLI and SST-2 datasets, following the methodology outlined in \cite{park2021vision}\footnote{This involves generating two orthogonal random vectors in a 1D flattened parameter space, which are then normalized and used to perturb the parameters. The strength of these perturbations is determined by their \(x\) and \(y\) coordinates, with the origin (0, 0) representing the unperturbed state of the parameters, and increasing distance indicating greater perturbation intensity.}. Our analysis reveals that SAFE yields a flatter loss landscape compared to LoRA. This flattening is attributed to SAFE's mechanism of controlling the norm of the weights through regularization effects (See Equation~\eqref{eq:regularization}), which consequently enhances resistance to weight perturbations, as depicted in Figure \ref{fig_7}(a).

\noindent
\textbf{Hessian Eigenvalue Spectrum Analysis.}
To quantitatively assess the visualized loss landscape shown in Figure \ref{fig_7}(a), we perform a detailed analysis of the top-5 Hessian eigenvalue spectrum. A pivotal finding in our analysis is the reduced magnitude of the maximum Hessian eigenvalue, which correlates with a flatter loss landscape, indicative of enhanced generalization potential. Moreover, the diminution of large Hessian eigenvalues facilitates more effective model training \cite{ghorbani2019investigation}. Furthermore, the suppression of the largest \textit{negative} Hessian eigenvalues markedly contributes to a more convex loss landscape, enhancing the stability of the training process. Figure \ref{fig_7}(b) demonstrates that SAFE not only effectively reduces the magnitude of Hessian eigenvalues relative to LoRA but also leads to a smoother and more consistent loss landscape. This evidence highlights the advantages of SAFE in promoting a more reliable and steady training behavior for adapter-tuning.

\subsection{Resource Efficiency} \label{Section 5.4}
We evaluate the resource efficiency of SAFE in terms of memory usage, computation amount, and training time. Table~\ref{table_4} shows the average resource efficiency improvement for the main results on NLU, QA and NLG tasks. Overall, SAFE reduces memory usage, computation amount, and training time by 42.85\%, 34.59\%, and 11.82\% compared to LoRA, respectively (on average). This means that SAFE can fine-tune twice as many downstream tasks under the same FLOPs budget and further enable on-device fine-tuning for personalization. For example, when fine-tuning a $\text{RoBERTa}_\text{large}$ model with a question answering downstream task, SAFE reduces memory usage from 17.73GB to 3.56GB; 8GB is the usual memory size of the edge devices.

\subsection{Expanded Experimental Results}
\textbf{Image Classification Task Evaluations.}
In Appendix \ref{Results on Image Classification Tasks}, we conduct comprehensive evaluations of SAFE on a variety of image classification tasks. These experiments consistently demonstrate the efficacy of SAFE, confirming its robust performance across diverse vision-related applications.\\
\textbf{Compatibility with Advanced Adapters.}
Further discussions on the integration of SAFE with various advanced adapter modules are presented in Appendix \ref{Results with Various PEFT Methods}. Our results highlight SAFE's versatility and compatibility with multiple adapter-tuning frameworks~\cite{houlsby2019parameter, zaken2022bitfit}. This adaptability ensures that SAFE's methodology remains effective, independent of specific adapter designs, thereby facilitating scalability across existing adapter-tuning methods.

%%%%%%%%%%%%%%%%%%%%%%%%%%%%%%%%%%%%%%%%%%%%%%%%%%%%%%%%%%
% TABLE_START
%%%%%%%%%%%%%%%%%%%%%%%%%%%%%%%%%%%%%%%%%%%%%%%%%%%%%%%%%%
\begin{table}
\centering
% \vskip -0.1in
\renewcommand{\arraystretch}{1.1}
\setlength\tabcolsep{1.5pt}
\captionof{table}{SAFE improves efficiency over the baseline in all aspects including memory usage, computation amount and training time. Note that we report computation costs for 1-step training.}
\label{table_4}
% \scriptsize
\fontsize{8}{9}\selectfont 
% \begin{center}
\begin{tabular}{@{}cl|c|c|c@{}}
\toprule
\multicolumn{1}{l}{}                        
    &          
    % & CV    
    & NLU         
    & QA     
    & NLG \\ \bottomrule
Memory Usage          
    & LoRA 
        % & 12.82 
        & 20.35        
        & 12.75  
        & 16.97    \\
(GB)
    & \cellcolor[HTML]{F2F2F2}+ SAFE    
        % & \cellcolor[HTML]{F2F2F2}8.16  {\fontsize{5pt}{6pt}\selectfont\textcolor{blue}{36.36\%$\downarrow$}} 
        & \cellcolor[HTML]{F2F2F2}12.11 {\fontsize{5pt}{6pt}\selectfont\textcolor{blue}{40.47\%$\downarrow$}} 
        & \cellcolor[HTML]{F2F2F2}5.85 {\fontsize{5pt}{6pt}\selectfont\textcolor{blue}{54.08\%$\downarrow$}} 
        & \cellcolor[HTML]{F2F2F2}11.90 {\fontsize{5pt}{6pt}\selectfont\textcolor{blue}{29.91\%$\downarrow$}}    \\ \midrule
Computational Cost                     
    & LoRA
        % & 8.32
        & 1.24            
        & 5.94   
        & 8.64    \\
(TFLOPs)
    & \cellcolor[HTML]{F2F2F2}+ SAFE    
        % & \cellcolor[HTML]{F2F2F2}6.43 {\fontsize{5pt}{6pt}\selectfont\textcolor{blue}{22.75\%$\downarrow$}}
        & \cellcolor[HTML]{F2F2F2}0.93 {\fontsize{5pt}{6pt}\selectfont\textcolor{blue}{24.74\%$\downarrow$}}      
        & \cellcolor[HTML]{F2F2F2}2.33 {\fontsize{5pt}{6pt}\selectfont\textcolor{blue}{60.86\%$\downarrow$}}  
        & \cellcolor[HTML]{F2F2F2}6.79 {\fontsize{5pt}{6pt}\selectfont\textcolor{blue}{21.42\%$\downarrow$}}    \\ \midrule
Training Time 
    & LoRA
        % & 1     
        & 1           
        & 1      
        & 1   \\
(Normalized)  
    & \cellcolor[HTML]{F2F2F2}+ SAFE    
        % & \cellcolor[HTML]{F2F2F2}0.88 {\fontsize{5pt}{6pt}\selectfont\textcolor{blue}{12.35\%$\downarrow$}}
        & \cellcolor[HTML]{F2F2F2}0.90 {\fontsize{5pt}{6pt}\selectfont\textcolor{blue}{10.29\%$\downarrow$}}
        & \cellcolor[HTML]{F2F2F2}0.89 {\fontsize{5pt}{6pt}\selectfont\textcolor{blue}{10.92\%$\downarrow$}}
        & \cellcolor[HTML]{F2F2F2}0.80 {\fontsize{5pt}{6pt}\selectfont\textcolor{blue}{19.76\%$\downarrow$}}   \\ \bottomrule
\end{tabular}
% \end{center}
% \vskip -0.1in
\end{table}
\section{Conclusion}
In this paper, we propose SAFE, which selectively freezes adapters for enabling resource efficient fine-tuning of PLMs. We observe that not all adapters contribute equally to adaptation. Motivated by the observation, SAFE gradually freezes less-important adapters, which do not contribute to adaptation during the early training steps. In our evaluation on various models and datasets, SAFE significantly saves memory usage and computation and accelerating training time, with comparable (or even better) accuracy. We also demonstrate that SAFE induces regularization effect, thereby improving generalization performance and accuracy compared to the state-of-the-art PEFT methods. We believe that SAFE can enable resource-efficient fine-tuning of large-scale PLMs, and further pave the path forward to personalized fine-tuning on resource-constrained edge devices.

% We observe that although adapter-tuning inserts the same number of trainable parameters into each attention layer, the impact of each adapter on resource usage and accuracy is not the same. In this paper, we propose SAFE, which reduces resource usage by early freezing of adapters that are not important for adaptation based on the importance score. Our extensive experiments show that SAFE automatically freezes adapters, significantly saving memory usage and computation, and accelerating training time, without compromising accuracy. We highlight that the regularization effect of selective early freezing has the advantage of loss landscape smoothing, preventing accuracy degradation, thus achieving a superior trade-off between accuracy and resource usage compared to baseline. We believe that the proposed method can be a resource-efficient fine-tuning solution for large-scale PLMs since it does not require any hardware modifications.

%%%%%%%%%%%%%%%%%%%%%%%%%%%%%%%%%%%%%%%%%%%%%%%%%%%%%%%%%
% Limitations
%%%%%%%%%%%%%%%%%%%%%%%%%%%%%%%%%%%%%%%%%%%%%%%%%%%%%%%%%
\section{Limitations}
We suggest the need for combination with prior research on memory-efficient training. These include low precision, micro-batching, weight sharding, and gradient checkpointing techniques. Though we have not evaluated SAFE along with such memory-efficient training methods, SAFE can be complementarily used along with the methods since SAFE can be applied independently of the training method or weight precision. In particular, since the quantization-based compression technique is quite popular and effective in terms of both compression ratio and preservation of final accuracy, favorable results are expected from combining the proposed technique with the memory-efficient training methods \cite{han2015deep}.

% Our research primarily focuses on reducing the memory requirement of fine-tuning large-scale PLMs. The carbon emissions produced by large-scale models fine-tuning may pose environmental issues. Our next step is to further improve the efficiency of fine-tuning large-scale pre-trained models, particularly on hardware with greater resource constraints.

\section*{Acknowledgments}
This work was supported in part by National Research Foundation of Korea (NRF) grant funded by the
Korea government (MSIT) (RS-2023-00212711), Institute of Information \& Communications Technology Planning \& Evaluation(IITP)-ITRC(Information Technology Research Center) grant funded by the MSIT(IITP-2025-RS-2023-00260091), ICT Creative Consilience Program through IITP grant funded by the MSIT(IITP-2025-RS-2020-II201819), IITP grant funded by the MSIT (No. RS-2024-00398353, Development of Countermeasure Technologies for Generative AI Security Threats). We also thank the members of the Korea University Intelligent Computer Architecture \& Systems research Lab for their useful comments and discussions, as well as the anonymous reviewers for their helpful feedback.

% \clearpage
\bibliography{coling_latex}

\clearpage
%%%%%%%%%%%%%%%%%%%%%%%%%%%%%%%%%%%%%%%%%%%%%%%%%%%%%%%%%
% APPENDIX
%%%%%%%%%%%%%%%%%%%%%%%%%%%%%%%%%%%%%%%%%%%%%%%%%%%%%%%%
\appendix
\onecolumn

\section{Results on Image Classification Tasks} \label{Results on Image Classification Tasks}
We conduct experiments with 8 datasets including class-level transfer and task-level transfer in the image classification tasks within Computer Vision (CV) domain. These datasets include CIFAR-10, CIFAR-100 \cite{krizhevsky2009learning}, Country-211 \cite{radford2021learning}, Fashion MNIST \cite{xiao2017fashion}, Food-101 \cite{bossard2014food}, Oxford Flowers \cite{nilsback2008automated}, Standford Cars \cite{krause20133d} and Tiny ImageNet \cite{le2015tiny}. 

Table \ref{table_5} shows that SAFE can effectively reduce memory usage while achieving comparable accuracy on eight datasets in the image classification task. For example, SAFE achieves a 50.69\% memory usage reduction on $\text{ViT}_\text{large}$ while maintaining comparable accuracy. Additionally, SAFE remains consistently effective regardless of variations in the pre-trained model size and backbone structure.

%%%%%%%%%%%%%%%%%%%%%%%%%%%%%%%%%%%%%%%%%%%%%%%%%%%%%%%%%%
% TABLE_BEGIN
%%%%%%%%%%%%%%%%%%%%%%%%%%%%%%%%%%%%%%%%%%%%%%%%%%%%%%%%%%
\begin{table*}[ht]
% \vskip -0.2in
\centering
\renewcommand{\arraystretch}{1.1}
\caption{Experimental results on eight common computer vision tasks. SAFE significantly reduces memory usage while achieving accuracy comparable to the baseline. Note that \textcolor{blue}{blue} indicates the memory usage reduction rate of SAFE compared to the baseline.}
\label{table_5}
% \scriptsize
\fontsize{8}{9}\selectfont 
% \tiny
\setlength\tabcolsep{1.5pt}
\begin{tabular}{@{}cl|cl|cl|cl|cl|cl|cl|cl|cl|cl@{}}
\toprule
 & 
 & 
  \multicolumn{2}{c|}{\fontsize{7.5pt}{6pt}\selectfont{CIFAR-10}} &
  \multicolumn{2}{c|}{\fontsize{7.5pt}{6pt}\selectfont{CIFAR-100}} &
  \multicolumn{2}{c|}{\fontsize{7.5pt}{6pt}\selectfont{Country-211}} &
  \multicolumn{2}{c|}{\fontsize{7.5pt}{6pt}\selectfont{Fashion MNIST}} &
  \multicolumn{2}{c|}{\fontsize{7.5pt}{6pt}\selectfont{Food-101}} &
  \multicolumn{2}{c|}{\fontsize{7.5pt}{6pt}\selectfont{Oxford Flowers}} &
  \multicolumn{2}{c|}{\fontsize{7.5pt}{6pt}\selectfont{Stanford Cars}} &
  \multicolumn{2}{c|}{\fontsize{7.5pt}{6pt}\selectfont{Tiny ImageNet}} &
  \multicolumn{2}{c}{Avg.} \\ \bottomrule
 &
  LoRA &
  \multicolumn{2}{c|}{97.77} &
  \multicolumn{2}{c|}{95.68} &
  \multicolumn{2}{c|}{16.56} &
  \multicolumn{2}{c|}{94.52} &
  \multicolumn{2}{c|}{88.84} &
  \multicolumn{2}{c|}{99.41} &
  \multicolumn{2}{c|}{82.56} &
  \multicolumn{2}{c|}{88.90} &
  \multicolumn{2}{c}{\textbf{83.03}} \\
\multirow{-2}{*}{$\text{ViT}_{\text{base}}$} &
  \cellcolor[HTML]{F2F2F2}+ SAFE &
  \multicolumn{2}{c|}{\cellcolor[HTML]{F2F2F2}98.66} &
  \multicolumn{2}{c|}{\cellcolor[HTML]{F2F2F2}95.55} &
  \multicolumn{2}{c|}{\cellcolor[HTML]{F2F2F2}16.29} &
  \multicolumn{2}{c|}{\cellcolor[HTML]{F2F2F2}93.71} &
  \multicolumn{2}{c|}{\cellcolor[HTML]{F2F2F2}88.78} &
  \multicolumn{2}{c|}{\cellcolor[HTML]{F2F2F2}99.61} &
  \multicolumn{2}{c|}{\cellcolor[HTML]{F2F2F2}82.05} &
  \multicolumn{2}{c|}{\cellcolor[HTML]{F2F2F2}89.16} &
  \multicolumn{2}{c}{\cellcolor[HTML]{F2F2F2}82.98 \textbf{{\fontsize{5pt}{6pt}\selectfont\textcolor{blue}{24.35\%$\downarrow$}}}} \\ \midrule
 &
  LoRA &
  \multicolumn{2}{c|}{99.09} &
  \multicolumn{2}{c|}{96.71} &
  \multicolumn{2}{c|}{20.44} &
  \multicolumn{2}{c|}{94.92} &
  \multicolumn{2}{c|}{90.32} &
  \multicolumn{2}{c|}{-} &
  \multicolumn{2}{c|}{86.97} &
  \multicolumn{2}{c|}{92.13} &
  \multicolumn{2}{c}{82.94} \\
\multirow{-2}{*}{$\text{ViT}_{\text{large}}$} &
  \cellcolor[HTML]{F2F2F2}+ LoRA &
  \multicolumn{2}{c|}{\cellcolor[HTML]{F2F2F2}99.13} &
  \multicolumn{2}{c|}{\cellcolor[HTML]{F2F2F2}97.00} &
  \multicolumn{2}{c|}{\cellcolor[HTML]{F2F2F2}20.44} &
  \multicolumn{2}{c|}{\cellcolor[HTML]{F2F2F2}94.88} &
  \multicolumn{2}{c|}{\cellcolor[HTML]{F2F2F2}90.69} &
  \multicolumn{2}{c|}{\cellcolor[HTML]{F2F2F2}-} &
  \multicolumn{2}{c|}{\cellcolor[HTML]{F2F2F2}86.83} &
  \multicolumn{2}{c|}{\cellcolor[HTML]{F2F2F2}92.03} &
  \multicolumn{2}{c}{\cellcolor[HTML]{F2F2F2}\textbf{83.00 {\fontsize{5pt}{6pt}\selectfont\textcolor{blue}{50.69\%$\downarrow$}}}} \\ \midrule
 &
  LoRA &
  \multicolumn{2}{c|}{98.92} &
  \multicolumn{2}{c|}{96.20} &
  \multicolumn{2}{c|}{20.06} &
  \multicolumn{2}{c|}{95.12} &
  \multicolumn{2}{c|}{91.36} &
  \multicolumn{2}{c|}{99.50} &
  \multicolumn{2}{c|}{87.14} &
  \multicolumn{2}{c|}{90.31} &
  \multicolumn{2}{c}{84.83} \\
\multirow{-2}{*}{$\text{SWIN}_{\text{base}}$} &
  \cellcolor[HTML]{F2F2F2}+ SAFE &
  \multicolumn{2}{c|}{\cellcolor[HTML]{F2F2F2}98.94} &
  \multicolumn{2}{c|}{\cellcolor[HTML]{F2F2F2}96.31} &
  \multicolumn{2}{c|}{\cellcolor[HTML]{F2F2F2}20.26} &
  \multicolumn{2}{c|}{\cellcolor[HTML]{F2F2F2}95.12} &
  \multicolumn{2}{c|}{\cellcolor[HTML]{F2F2F2}91.51} &
  \multicolumn{2}{c|}{\cellcolor[HTML]{F2F2F2}99.71} &
  \multicolumn{2}{c|}{\cellcolor[HTML]{F2F2F2}86.88} &
  \multicolumn{2}{c|}{\cellcolor[HTML]{F2F2F2}90.14} &
  \multicolumn{2}{c}{\cellcolor[HTML]{F2F2F2}\textbf{84.86 {\fontsize{5pt}{6pt}\selectfont\textcolor{blue}{32.01\%$\downarrow$}}}} \\ \midrule
 &
  LoRA &
  \multicolumn{2}{c|}{99.07} &
  \multicolumn{2}{c|}{97.01} &
  \multicolumn{2}{c|}{22.19} &
  \multicolumn{2}{c|}{95.44} &
  \multicolumn{2}{c|}{92.68} &
  \multicolumn{2}{c|}{-} &
  \multicolumn{2}{c|}{85.06} &
  \multicolumn{2}{c|}{92.02} &
  \multicolumn{2}{c}{83.35} \\
\multirow{-2}{*}{$\text{SWIN}_{\text{large}}$} &
  \cellcolor[HTML]{F2F2F2}+ LoRA &
  \multicolumn{2}{c|}{\cellcolor[HTML]{F2F2F2}99.17} &
  \multicolumn{2}{c|}{\cellcolor[HTML]{F2F2F2}96.72} &
  \multicolumn{2}{c|}{\cellcolor[HTML]{F2F2F2}22.66} &
  \multicolumn{2}{c|}{\cellcolor[HTML]{F2F2F2}95.47} &
  \multicolumn{2}{c|}{\cellcolor[HTML]{F2F2F2}92.69} &
  \multicolumn{2}{c|}{\cellcolor[HTML]{F2F2F2}-} &
  \multicolumn{2}{c|}{\cellcolor[HTML]{F2F2F2}85.13} &
  \multicolumn{2}{c|}{\cellcolor[HTML]{F2F2F2}92.11} &
  \multicolumn{2}{c}{\cellcolor[HTML]{F2F2F2}\textbf{83.42 {\fontsize{5pt}{6pt}\selectfont\textcolor{blue}{25.06\%$\downarrow$}}}} \\ \bottomrule
\end{tabular}
\end{table*}
%%%%%%%%%%%%%%%%%%%%%%%%%%%%%%%%%%%%%%%%%%%%%%%%%%%%%%%%%%
% TABLE_END
%%%%%%%%%%%%%%%%%%%%%%%%%%%%%%%%%%%%%%%%%%%%%%%%%%%%%%%%%%

\section{Results with Various PEFT Methods} \label{Results with Various PEFT Methods}
We validate the applicability of SAFE upon advanced adapter modules \cite{houlsby2019parameter, zaken2022bitfit, hu2021lora}. Table \ref{table_6} shows that SAFE reduces memory usage by 24.76\% on average while achieving a comparable GLUE score. This result demonstrates that SAFE can be applied to a variety of adapter-tuning methods to enable resource efficient fine-tuning of large language models.
%%%%%%%%%%%%%%%%%%%%%%%%%%%%%%%%%%%%%%%%%%%%%%%%%%%%%%%%%%
% TABLE_START
%%%%%%%%%%%%%%%%%%%%%%%%%%%%%%%%%%%%%%%%%%%%%%%%%%%%%%%%%%
\begin{table*}[ht]
% \vskip -0.2in
\centering
\renewcommand{\arraystretch}{1.5}
\caption{Experimental results for various adapter-tuning methods on the GLUE benchmark. Note that blue indicates the memory usage reduction rate of SAFE compared to the baseline.}
\label{table_6}
% \scriptsize
% \tiny
\fontsize{8}{9}\selectfont 
\setlength\tabcolsep{1.5pt}
\begin{tabular}{@{}l|cccccccc|c@{}}
\toprule
 & CoLA  & SST-2 & MNLI  & RTE   & QQP   & MRPC  & QNLI  & STS-B &       Avg.      
 \\  \multicolumn{1}{c|}{$\text{BERT}_\text{base}$}
   &    Matthews corr.
   &    Accuracy
   &    Accuracy
   &    Accuracy
   &    Accuracy
   &    F1 Score
   &    Accuracy
   &    Pearson corr.                          
   &    \\ 
 \bottomrule
  % \multicolumn{10}{c}{\textbf{\textit{Baselines}}}                                                       \\ \hline
  Houlsby  
    & 62.38 & 91.17 & 83.44 & 70.50 & 90.85 & 89.52 & 90.59 & 90.75 & 83.65 \\
  \cellcolor[HTML]{F2F2F2}{+ SAFE} 
  & \cellcolor[HTML]{F2F2F2}62.83 
  & \cellcolor[HTML]{F2F2F2}93.00 
  & \cellcolor[HTML]{F2F2F2}84.15 
  & \cellcolor[HTML]{F2F2F2}74.10 
  & \cellcolor[HTML]{F2F2F2}90.93 
  & \cellcolor[HTML]{F2F2F2}89.70 
  & \cellcolor[HTML]{F2F2F2}91.03 
  & \cellcolor[HTML]{F2F2F2}91.29 
  & \cellcolor[HTML]{F2F2F2}\textbf{84.63{\fontsize{5pt}{6pt}\selectfont\textcolor{blue}{25.70\%$\downarrow$}}} \\ \hline

  BitFit 
    & 60.92 & 91.86 & 82.41 & 71.94 & 89.20 & 88.14 & 89.80 & 90.95 & \textbf{83.15} \\
  \cellcolor[HTML]{F2F2F2}{+ SAFE}  
  & \cellcolor[HTML]{F2F2F2}61.73 
  & \cellcolor[HTML]{F2F2F2}93.00 
  & \cellcolor[HTML]{F2F2F2}81.85 
  & \cellcolor[HTML]{F2F2F2}69.78 
  & \cellcolor[HTML]{F2F2F2}89.21 
  & \cellcolor[HTML]{F2F2F2}88.85 
  & \cellcolor[HTML]{F2F2F2}89.51 
  & \cellcolor[HTML]{F2F2F2}90.75 
  & \cellcolor[HTML]{F2F2F2}83.09 \textbf{{\fontsize{5pt}{6pt}\selectfont\textcolor{blue}{25.53\%$\downarrow$}}}\\ \hline
  LoRA
   & 64.46 & 91.63 & 82.88 & 71.22 & 90.01 & 88.39 & 90.01 & 90.86 & 83.68 \\
 \cellcolor[HTML]{F2F2F2}{ + SAFE} &
  \cellcolor[HTML]{F2F2F2}66.80 &
  \cellcolor[HTML]{F2F2F2}90.83 &
  \cellcolor[HTML]{F2F2F2}82.03 &
  \cellcolor[HTML]{F2F2F2}71.22 &
  \cellcolor[HTML]{F2F2F2}89.74 &
  \cellcolor[HTML]{F2F2F2}88.51 &
  \cellcolor[HTML]{F2F2F2}90.65 &
  \cellcolor[HTML]{F2F2F2}90.26  &
  \cellcolor[HTML]{F2F2F2}\textbf{83.76{\fontsize{5pt}{6pt}\selectfont\textcolor{blue}{23.06\%$\downarrow$}}} \\ \bottomrule 
\end{tabular}
\end{table*}

%%%%%%%%%%%%%%%%%%%%%%%%%%%%%%%%%%%%%%%%%%%%%%%%%%%%%%%%%%
% TABLE_END
%%%%%%%%%%%%%%%%%%%%%%%%%%%%%%%%%%%%%%%%%%%%%%%%%%%%%%%%%%

\clearpage
\section{Experimental Setup} \label{Experimental Setups}
% \subsection{Adapter Budget Scheduler}

% Let $F_t$ be the set of frozen layers at epoch t and $N$ is total number of training epochs The freezing stage, $S$ is defined as:

% \begin{align} \label{equation_6}
% % \vskip -0.2in
% S = \min \{ t \mid F_t = F_{t+1}, 0 < t < N \times  \frac{n}{100}\}
% % \vskip -0.2in
% \end{align}

\subsection{Model}
We conduct experiments using a pre-trained model deployed on HuggingFace \cite{wolf2019huggingface}. For experiments on the NLU and QA benchmarks, we use bert-base-uncased and bert-large-uncased trained on BookCorpus, a dataset consisting of 11,038 unpublished books and English Wikipedia. We use roberta-base and roberta-large trained on 5 datasets (BookCorpus, English Wikipedia, CC-News, OpenWebText, and Stories) for the RoBERTa model. For experiments on the NLG benchmark, we use GPT-2 medium and GPT-2 large distributed by OpenAI. We also employ the LLaMA-2 7B model, an open-weight large language model released by Meta, which has been trained on a mixture of publicly available and licensed datasets, including Common Crawl, C4, GitHub, Wikipedia, Project Gutenberg, ArXiv, and Stack Exchange. We use vit-base-patch16-224-in21k and vit-large-patch16-224-in21k distributed by Google for experiments in the ViT model. Finally, We use the swin-base-patch4-window7-224 and swin-large-patch4-window7-224 models distributed by Microsoft for experiments in the SWIN model.

\subsection{Computing Resources} 
Our experimental setup leverages 2 RTX4090 with 24GB memory for NLU, QA, and NLG tasks and 1 RTX 4090 for CV downstream task, excluding the LLaMA experiments.

\subsection{Dataset Statistics} \label{Dataset Statistics}
We present the dataset statistics of GLUE~\cite{wang2018glue}, SQuAD~\cite{rajpurkar2016squad}, E2E NLG Challenge~\cite{novikova2017e2e}, and WikiText-2~\cite{merity2022pointer} in following table. We fine-tune models on the LaMini instruction dataset~\cite{wu2024lamini} for \( \text{LLaMA-2}_{\text{7B}} \) and evaluate their performance on WikiText-2 using perplexity.

\begin{table*}[ht]
%\vskip -0.2in
\caption{Summary of the NLU, QA, and NLG benchmarks.}
\label{table_9}
% \scriptsize
% \footnotesize
\fontsize{8.5}{10}\selectfont 
\setlength\tabcolsep{2pt}
\renewcommand{\arraystretch}{1.1}
% \scriptsize % This will make the font smaller; you can also try \scriptsize if it's still too large
\centering
\begin{tabular}{@{}lrrrrll@{}}
\toprule
\multicolumn{7}{c}{\textbf{NLU Benchmark}}                                         \\ \midrule
\multicolumn{1}{c}{Dataset} &
  \multicolumn{1}{c}{\# Train} &
  \multicolumn{1}{c}{\# Valid} &
  \multicolumn{1}{c}{\# Test} &
  \multicolumn{1}{c}{\# Label} &
  \multicolumn{1}{c}{Task} &
  \multicolumn{1}{c}{Evaluation Metric} \\ \bottomrule
\multicolumn{7}{c}{Single-Sentence Classification (GLUE)}       \\ \midrule
CoLA  & 8,551   & 521   & 522    & 2 & Acceptability      & Matthews corr \\
SST-2 & 66,349  & 1,000 & 872    & 2 & Sentiment          & Accuracy      \\ \midrule
\multicolumn{7}{c}{Pairwise Text Classification (GLUE)}         \\ \midrule
MNLI  & 392,702 & 9,832 & 9,815  & 3 & NLI                & Accuracy      \\
RTE   & 2,490   & 138   & 139    & 2 & NLI                & Accuracy      \\
QQP   & 362,846 & 1,000 & 40,431 & 2 & Paraphrase         & Accuracy      \\
MRPC  & 3,668   & 204   & 204    & 2 & Paraphrase         & F1 score      \\
QNLI  & 103,743 & 1,000 & 5,463  & 2 & QA/NLI             & Accuracy      \\ \midrule
\multicolumn{7}{c}{Pairwise Text Classification (GLUE)}         \\ \midrule
STS-B & 5,749   & 750   & 750    & 1 & Similarity         & Pearson corr  \\ \bottomrule
\multicolumn{7}{c}{\textbf{QA Benchmark}}                                   \\ \midrule
\multicolumn{1}{c}{Dataset} &
  \multicolumn{1}{c}{\# Train} &
  \multicolumn{1}{c}{\# Valid} &
  \multicolumn{1}{c}{\# Test} &
  \multicolumn{1}{c}{\# Label} &
  \multicolumn{1}{c}{Task} &
  \multicolumn{1}{c}{Evaluation Metric}                       \\ \bottomrule
SQuAD & 87,600  & 5,300 & 5,300  & 2 & Question Answering & F1 score      \\ \bottomrule
\multicolumn{7}{c}{\textbf{NLG Benchmark}}                                   \\ \midrule
\multicolumn{1}{c}{Dataset} &
  \multicolumn{1}{c}{\# Train} &
  \multicolumn{1}{c}{\# Valid} &
  \multicolumn{1}{c}{\# Test} &
  \multicolumn{1}{c}{\# Label} &
  \multicolumn{1}{c}{Task} &
  \multicolumn{1}{c}{Evaluation Metric}                       \\ \bottomrule
E2E NLG Challenge & 42,061  & 4,672 & 4,693  &   & Generation & BLEU, NIST, METEOR, ROUGE-L, and CIDEr      \\ 
WikiText-2 & 2,589,000  & 1,000 & 4,360  &   & Language Modeling & Perplexity (PPL) \\ \bottomrule
\end{tabular}
% \vskip -0.2in
\end{table*}

\clearpage
The following table lists dataset statistics evaluated in the CV domain.
\begin{table*}[ht]
% \vskip -0.2in
\caption{Summary of CV benchmark.}
\label{table_8}
\footnotesize
\renewcommand{\arraystretch}{1.2}
% \scriptsize % This will make the font smaller; you can also try \scriptsize if it's still too large
\centering
\begin{tabular}{@{}lrrrrcc@{}}
\toprule
\multicolumn{7}{c}{\textbf{CV Benchmark}}                         \\ \midrule
\multicolumn{1}{c}{Dataset} &
  \multicolumn{1}{c}{\# Train} &
  \multicolumn{1}{c}{\# Valid} &
  \multicolumn{1}{c}{\# Test} &
  \multicolumn{1}{c}{\# Label} &
  Task &
  Evaluation Metric \\ \bottomrule
CIFAR-10       & 45,000 & 5,000 & 10,000 & 10  & Classification & Accuracy \\
CIFAR-100      & 45,000 & 5,000 & 10,000 & 100 & Classification & Accuracy \\
Fashion MNIST  & 54,000 & 6,000 & 10,000 & 10  & Classification & Accuracy \\
Oxford Flowers & 6,453  & 717   & 1,020  & 102 & Classification & Accuracy \\
Food-101       & 68,220 & 7,580 & 25,300 & 102 & Classification & Accuracy \\
Country-211    & 25,920 & 2,880 & 21,100 & 211 & Classification & Accuracy \\
Stanford Cars  & 7,326  & 814   & 8,040  & 196 & Classification & Accuracy \\
Tiny ImageNet  & 90,000 & 10,000& 10,000 & 200 & Classification & Accuracy \\ \bottomrule
\end{tabular}
% \vskip -0.2in
\end{table*}

\newpage
\subsection{Hyperparameter Settings} \label{Hyperparameter Settings}
We explore 10\% of all epochs for at least 5 learning rates. Hyperparameter settings, including learning rate, are made by referring to previous works \cite{he2023parameter, houlsby2019parameter, hu2021lora, zaken2022bitfit}. We use the AdamW optimizer \cite{loshchilov2018decoupled} and LinearLR learning rate scheduler and set weight decay to 0 in experiments. In our evaluation, we configure LoRA as follows: \(r\) = 4, alpha = 16, target modules = ["query", "value"], and LoRA dropout = 0.1. 

Since the lower layers of the BERT model are more easily converted to fine-tuning than the upper layers, we extend the previous study~\cite{zhangrevisiting} that froze most of the lower layers with adapter-tuning and experiment with a baseline that naively inserts adapters only into six layers. For LoRAPrune, we configure the pruning ratio to 0.5, pruning frequency to 10, and the pruning metric to 'lora,' as these settings have shown effective results in prior work \cite{zhang2023loraprune}. In the case of MEFT, for a fair comparison, we set the reversible layers to half of the total number of layers. We employ the $\text{MEFT}_{1}$ method, where the $\mathcal{F}$ architecture is the layer, from the three reversible transformation methods proposed in MEFT. For the hyperparameters of MEFT, such as \textit{$\lambda$} and \textit{$\beta$}, we follow the values proposed in previous work~\cite{liao2024make} to ensure consistency in the evaluation.

\begin{table*}[ht]
\caption{Hyperparameter settings on the NLU and QA tasks.}
\label{table_11}
% \small 
\renewcommand{\arraystretch}{1.2}
\scriptsize % This will make the font smaller; you can also try \scriptsize if it's still too large
\centering
\begin{tabular}{@{}cllccc@{}}
\toprule
pre-trained model                                & \multicolumn{1}{c}{dataset}  & \multicolumn{1}{c}{method} & final   learning rate & batch   size & \# epochs \\ \bottomrule
\multirow{27}{*}{\textbf{BERT-base-uncased}}  & \multirow{3}{*}{GLUE / CoLA} & LoRA, LoRA + SAFE          & 6.00E-04              & 32           & 100       \\
              &                               & BitFit, BitFit + SAFE     & 9.00E-04 & 32 & 100 \\
              &                               & Houlsby, Houlsby + SAFE & 5.00E-04 & 32 & 100 \\
              & \multirow{3}{*}{GLUE / SST-2} & LoRA, LoRA + SAFE         & 7.00E-04 & 32 & 75  \\
              &                               & BitFit, BitFit + SAFE     & 7.00E-04 & 32 & 75  \\
              &                               & Houlsby, Houlsby + SAFE & 2.00E-04 & 32 & 75  \\
              & \multirow{3}{*}{GLUE / MNLI}  & LoRA, LoRA + SAFE         & 9.00E-04 & 32 & 50  \\
              &                               & BitFit, BitFit + SAFE     & 8.00E-04 & 32 & 50  \\
              &                               & Houlsby, Houlsby + SAFE & 4.00E-04 & 32 & 50  \\
              & \multirow{3}{*}{GLUE / RTE}   & LoRA, LoRA + SAFE         & 9.00E-04 & 32 & 100 \\
              &                               & BitFit, BitFit + SAFE     & 8.00E-04 & 32 & 100 \\
              &                               & Houlsby, Houlsby + SAFE & 4.00E-04 & 32 & 100 \\
              & \multirow{3}{*}{GLUE / QQP}   & LoRA, LoRA + SAFE         & 4.00E-04 & 32 & 50  \\
              &                               & BitFit, BitFit + SAFE     & 6.00E-04 & 32 & 50  \\
              &                               & Houlsby, Houlsby + SAFE & 4.00E-04 & 32 & 50  \\
              & \multirow{3}{*}{GLUE / MRPC}  & LoRA, LoRA + SAFE         & 5.00E-04 & 16 & 50  \\
              &                               & BitFit, BitFit + SAFE     & 5.00E-04 & 32 & 50  \\
              &                               & Houlsby, Houlsby + SAFE & 5.00E-04 & 32 & 50  \\
              & \multirow{3}{*}{GLUE / QNLI}  & LoRA, LoRA + SAFE         & 5.00E-04 & 32 & 50  \\
              &                               & BitFit, BitFit + SAFE     & 5.00E-04 & 32 & 50  \\
              &                               & Houlsby, Houlsby + SAFE & 4.00E-04 & 32 & 50  \\
              & \multirow{3}{*}{GLUE / STS-B} & LoRA, LoRA + SAFE         & 8.00E-04 & 32 & 50  \\
              &                               & BitFit, BitFit + SAFE     & 9.00E-04 & 32 & 50  \\
              &                               & Houlsby, Houlsby + SAFE & 5.00E-04 & 32 & 50  \\
              & \multirow{3}{*}{SQuAD}        & LoRA, LoRA + SAFE         & 3.00E-04 & 16 & 50  \\
              &                               & BitFit, BitFit + SAFE     & 9.00E-04 & 16 & 50  \\
              &                               & Houlsby, Houlsby + SAFE & 1.00E-04 & 16 & 50  \\ \midrule
\multirow{14}{*}{\textbf{BERT-large-uncased}} & GLUE / CoLA                  & LoRA, LoRA + SAFE          & 1.00E-04& 32           & 80        \\
              & GLUE / SST-2                  & LoRA, LoRA + SAFE         & 6.00E-04 & 32 & 60  \\
              & GLUE / MNLI                   & LoRA, LoRA + SAFE         & 1.00E-04 & 16 & 40  \\
              & GLUE / RTE                    & LoRA, LoRA + SAFE         & 6.00E-04 & 32 & 80  \\
              & GLUE / QQP                    & LoRA, LoRA + SAFE         & 3.00E-04 & 16 & 40  \\
              & GLUE / MRPC                   & LoRA, LoRA + SAFE         & 3.00E-04 & 4  & 50  \\
              & GLUE / QNLI                   & LoRA, LoRA + SAFE         & 2.00E-04& 8& 50  \\
              & GLUE / STS-B                  & LoRA, LoRA + SAFE         & 8.00E-04 & 32 & 50  \\
              & \multirow{6}{*}{SQuAD}        & Full-param   Fine-tuning  & 7.00E-05 & 16 & 50  \\
              &                               & LoRA,   LoRA + SAFE       & 3.00E-04 & 16 & 50  \\
              &                               & BitFit, BitFit + SAFE     & 9.00E-04 & 16 & 50  \\
              &                               & Houlsby, Houlsby + SAFE   & 1.00E-04 & 16 & 50  \\ 
              &                               & Pfeiffer, Pfeiffer + SAFE & 3.00E-04 & 16 & 50  \\ 
              &                               & AdaLoRA, AdaLoRA + SAFE   & 4.00E-04 & 16 & 50  \\
              % &                               & AdapterDrop       
              % & 3.00E-04 & 16 & 50 \\
              % &                               & SparseAdapter       
              % & ?.00E-04 & 16 & 50 \\
              \midrule
\multirow{3}{*}{\textbf{RoBERTa-base}}        
              %   & GLUE / CoLA                  & LoRA, LoRA + SAFE          & 8.00E-04              & 32           & 80        \\
              % & GLUE / SST-2                  & LoRA, LoRA + SAFE         & 7.00E-04 & 32 & 60  \\
              % & GLUE / MNLI                   & LoRA, LoRA + SAFE         & 9.00E-04 & 32 & 40  \\
              % & GLUE / RTE                    & LoRA, LoRA + SAFE         & 3.00E-04 & 32 & 80  \\
              % & GLUE / QQP                    & LoRA, LoRA + SAFE         & 4.00E-04 & 32 & 40  \\
              % & GLUE / MRPC                   & LoRA, LoRA + SAFE         & 6.00E-04 & 32 & 50  \\
              % & GLUE / QNLI                   & LoRA, LoRA + SAFE         & 3.00E-04 & 32 & 50  \\
              % & GLUE / STS-B                  & LoRA, LoRA + SAFE         & 3.00E-04 & 32 & 50  \\
 & \multirow{3}{*}{SQuAD}& LoRA, LoRA + SAFE         & 5.00E-04 & 32 &50  \\
 & & BitFit, BitFit+SAFE& 8.00E-04& 32 &50  \\
              & & Houlsby, Houlsby+SAFE& 4.00E-04& 32 & 50  \\ \midrule
\multirow{3}{*}{\textbf{RoBERTa-large}}        & \multirow{3}{*}{SQuAD}& LoRA, LoRA + SAFE         & 6.00E-04 & 16 & 50  \\ 
 & & BitFit, BitFit+SAFE
& 7.00E-04& 16 &50  \\
 & & Houlsby, Houlsby+SAFE& 4.00E-04& 16 &50  \\ \bottomrule
\end{tabular}
\end{table*}

% \begin{table*}[ht]
% \caption{Hyperparameter settings on the NLG task.}
% \label{table_12}
% % \small 
% \renewcommand{\arraystretch}{1.2}
% \scriptsize % This will make the font smaller; you can also try \scriptsize if it's still too large
% \centering
% \begin{tabular}{lllllllll}
%                             & \multicolumn{5}{c}{Training}  & \multicolumn{3}{c}{Inference} \\
%  & final learning rate & batch size & \# epochs & Seq Len & Label Smooth & Beam Size & Length Penalty & no repeat ngram size \\
% \multicolumn{1}{c}{\textbf{GPT-2 medium}} & 1.00E-04 & 8 & 10 & 512 & 0.1 & 10       & 0.8       & 4      \\
% \multicolumn{1}{c}{\textbf{GPT-2 large}} & 5.00E-05 & 4 & 10 & 512 & 0.1 & 10       & 0.8       & 4     
% \end{tabular}
% \end{table*}

\begin{table*}[ht]
\caption{Hyperparameter settings on the NLG task.}
\label{table_12}
% \small 
\renewcommand{\arraystretch}{1.2}
\footnotesize
% \scriptsize % This will make the font smaller; you can also try \scriptsize if it's still too large
\centering
\begin{tabular}{@{}lccc@{}}
\toprule
    pre-trained model
    & \multicolumn{1}{c}{\textbf{GPT-2 medium}} 
    & \multicolumn{1}{c}{\textbf{GPT-2 large}}
    % & \multicolumn{1}{c}{\textbf{GPT-2 xl}}
    & \multicolumn{1}{c}{\textbf{LLaMA-2 7B}}
    \\ \bottomrule
& \multicolumn{2}{c}{Training}                              \\ \hline
final learning rate     
    & 1.00E-04                    
    & 5.00E-05
    % & 
    & 1.00E-04
    \\
batch size              
    & 8                           
    & 4 
    % & 4 (w/ gradient accumulation) 
    & 128 (w/ micro batch size 2) 
    \\
\# epochs               
    & 10                          
    & 10
    % & 10 
    & 3
    \\
Seq Length              
    & 512                         
    & 512
    % & 512 
    & 512
    % \\
% Weight Decay            
%     & 0                           
%     & 0
%     % & 0 
%     \\
% Dropout Probability     
%     & 0.1                         
%     & 0.1
%     % & 0.1 
    \\
Label Smooth            
    & 0.1                         
    & 0.1
    % & 0.1 
    & 
    \\ \hline
& \multicolumn{2}{c}{Inference}                             \\ \hline
Beam Size               
    & 10                          
    & 10
    % & 10 
    &
    \\
Length Penalty          
    & 0.8                         
    & 0.8
    % & 0.8 
    &
    \\
no repeat ngram size    
    & 4                           
    & 4
    % & 4 
    &
    \\ \bottomrule
\end{tabular}
\end{table*}

\begin{table*}[ht]
\caption{Hyperparameter settings on the CV task.}
\label{table_10}
% \small
\renewcommand{\arraystretch}{1.2}
\scriptsize % This will make the font smaller; you can also try \scriptsize if it's still too large
\centering
\begin{tabular}{@{}cclccc@{}}
\toprule
pre-trained model                                                   & dataset                      & \multicolumn{1}{c}{method} & final   learning rate & batch   size & \# epochs \\ \bottomrule
\multirow{16}{*}{\textbf{ViT-base-patch16-224}}         & \multirow{2}{*}{CIFAR-10}    & BitFit, BitFit + SAFE      & 3.00E-03              & 64           & 100       \\
 &                                    & LoRA,   LoRA + SAFE   & 3.00E-03 & 64 & 100 \\
 & \multirow{2}{*}{CIFAR-100}         & BitFit, BitFit + SAFE & 3.00E-03 & 64 & 100 \\
 &                                    & LoRA,   LoRA + SAFE   & 3.00E-03 & 64 & 100 \\
 & \multirow{2}{*}{Fashion MNIST}     & BitFit, BitFit + SAFE & 4.00E-03 & 64 & 100 \\
 &                                    & LoRA,   LoRA + SAFE   & 3.00E-03 & 64 & 100 \\
 & \multirow{2}{*}{Oxford Flowers}    & BitFit, BitFit + SAFE & 2.00E-03& 64 & 30 \\
 &                                    & LoRA,   LoRA + SAFE   & 8.00E-04& 64 & 40  \\
 & \multirow{2}{*}{Food-101}          & BitFit, BitFit + SAFE & 4.00E-03 & 64 & 100 \\
 &                                    & LoRA,   LoRA + SAFE   & 3.00E-03 & 64 & 100 \\
 & \multirow{2}{*}{Tiny ImageNet}     & BitFit, BitFit + SAFE & 1.00E-03 & 64 & 100 \\
 &                                    & LoRA,   LoRA + SAFE   & 8.00E-04 & 64 & 100 \\
 & \multirow{2}{*}{Country-211}       & BitFit, BitFit + SAFE & 2.00E-03 & 64 & 100 \\
 &                                    & LoRA,   LoRA + SAFE   & 4.00E-03 & 64 & 100 \\
 & \multirow{2}{*}{Stanford Cars}     & BitFit, BitFit + SAFE & 9.00E-03 & 64 & 100 \\
 &                                    & LoRA,   LoRA + SAFE   & 7.00E-03 & 64 & 100 \\ \midrule
\multirow{16}{*}{\textbf{ViT-large-patch16-224}}        & \multirow{2}{*}{CIFAR-10}    & BitFit, BitFit + SAFE      & 4.00E-03              & 64           & 100       \\
 &                                    & LoRA,   LoRA + SAFE   & 6.00E-04 & 64 & 100 \\
 & \multirow{2}{*}{CIFAR-100}         & BitFit, BitFit + SAFE & 2.00E-03 & 64 & 100 \\
 &                                    & LoRA,   LoRA + SAFE   & 5.00E-04 & 64 & 100 \\
 & \multirow{2}{*}{Fashion MNIST}     & BitFit, BitFit + SAFE & 3.00E-03 & 64 & 100 \\
 &                                    & LoRA,   LoRA + SAFE   & 9.00E-04 & 64 & 100 \\
 & \multirow{2}{*}{Oxford Flowers}    & BitFit, BitFit + SAFE & -        & -  & -   \\
 &                                    & LoRA,   LoRA + SAFE   & -        & -  & -   \\
 & \multirow{2}{*}{Food-101}          & BitFit, BitFit + SAFE & 9.00E-04 & 64 & 100 \\
 &                                    & LoRA,   LoRA + SAFE   & 7.00E-04 & 64 & 100 \\
 & \multirow{2}{*}{Tiny ImageNet}     & BitFit, BitFit + SAFE & 8.00E-04 & 64 & 100 \\
 &                                    & LoRA,   LoRA + SAFE   & 6.00E-04 & 64 & 100 \\
 & \multirow{2}{*}{Country-211}       & BitFit, BitFit + SAFE & 2.00E-03 & 64 & 100 \\
 &                                    & LoRA,   LoRA + SAFE   & 9.00E-04 & 64 & 100 \\
 & \multirow{2}{*}{Stanford Cars}     & BitFit, BitFit + SAFE & 1.00E-03 & 64 & 100 \\
 &                                    & LoRA,   LoRA + SAFE   & 1.00E-03 & 64 & 100 \\ \midrule
\multirow{8}{*}{\textbf{SWIN-base-patch4-window7-224}}  & \multicolumn{1}{l}{CIFAR-10} & LoRA, LoRA + SAFE          & 1.00E-03              & 64           & 50        \\
 & \multicolumn{1}{l}{CIFAR-100}      & LoRA, LoRA + SAFE     & 1.00E-03 & 64 & 50  \\
 & \multicolumn{1}{l}{Fashion MNIST}  & LoRA, LoRA + SAFE     & 1.00E-03 & 64 & 50  \\
 & \multicolumn{1}{l}{Oxford Flowers} & LoRA, LoRA + SAFE     & 7.00E-04 & 64 & 30  \\
 & \multicolumn{1}{l}{Food-101}       & LoRA, LoRA + SAFE     & 9.00E-04 & 64 & 50  \\
 & \multicolumn{1}{l}{Tiny ImageNet}  & LoRA, LoRA + SAFE     & 1.00E-03 & 64 & 50  \\
 & \multicolumn{1}{l}{Country-211}    & LoRA, LoRA + SAFE     & 7.00E-04 & 64 & 50  \\
 & \multicolumn{1}{l}{Stanford Cars}  & LoRA, LoRA + SAFE     & 1.00E-03 & 64 & 50  \\ \midrule
\multirow{8}{*}{\textbf{SWIN-large-patch4-window7-224}} & \multicolumn{1}{l}{CIFAR-10} & LoRA, LoRA + SAFE          & 8.00E-04              & 64           & 50        \\
 & \multicolumn{1}{l}{CIFAR-100}      & LoRA, LoRA + SAFE     & 7.00E-04 & 64 & 50  \\
 & \multicolumn{1}{l}{Fashion MNIST}  & LoRA, LoRA + SAFE     & 1.00E-03 & 64 & 50  \\
 & \multicolumn{1}{l}{Oxford Flowers} & LoRA, LoRA + SAFE     & -        & -  & -   \\
 & \multicolumn{1}{l}{Food-101}       & LoRA, LoRA + SAFE     & 5.00E-04 & 64 & 50  \\
 & \multicolumn{1}{l}{Tiny ImageNet}  & LoRA, LoRA + SAFE     & 6.00E-04 & 64 & 50  \\
 & \multicolumn{1}{l}{Country-211}    & LoRA, LoRA + SAFE     & 6.00E-04 & 64 & 50  \\
 & \multicolumn{1}{l}{Stanford Cars}  & LoRA, LoRA + SAFE     & 3.00E-03 & 64 & 50  \\ \bottomrule
\end{tabular}
\end{table*}

\end{document}